\RequirePackage{fix-cm}

\documentclass[twocolumn]{svjour3}          

\smartqed  

\usepackage{url}
\usepackage{graphicx}
\usepackage{multirow}

\usepackage[dvipsnames]{xcolor} 
\usepackage{amsmath,amsfonts}   
\usepackage{amssymb}            
\usepackage{dsfont}             
\usepackage{microtype}          
\usepackage{array}              
\usepackage{enumitem}           
\usepackage{makecell}           
\usepackage{colortbl}           
\usepackage{tikz}               
\usepackage{algorithm}          
\usepackage{algpseudocode}      

\definecolor{Gray}{gray}{0.8}
\definecolor{lGray}{gray}{0.9}
\definecolor{featurebased}{rgb}{0.57, 0.66, 0.77}
\definecolor{rocket}{rgb}{0.78,0.15,0.95}
\definecolor{convbased}{rgb}{0.34, 0.59, 0.97}
\definecolor{transbased}{rgb}{0.95,0.71, 0.28}

\newcommand{\revone}[1]{\textcolor{black}{#1}}
\newcommand{\revtwo}[1]{\textcolor{black}{#1}}
\newcommand{\revthree}[1]{\textcolor{black}{#1}}

\newcommand{\trianglemarker}[1]{\tikz \fill[#1] (0,0) -- (0.22cm,0) -- (0.11cm,0.20cm) -- cycle;}
\newcommand{\xmarker}[1]{\tikz \draw[#1, thick] (0.08cm,0.08cm) -- (-0.08cm,-0.08cm) (0.08cm,-0.08cm) -- (-0.08cm,0.08cm);}
\newcommand{\circlemarker}[1]{\tikz \fill[#1] (0,0) circle (0.1cm);}
\newcommand{\rectanglemarker}[1]{\tikz \fill[#1] (0,0) rectangle (0.17cm,0.17cm);}

\newcommand\vldbavailabilityurl{https://github.com/sylligardos/ff-vus}

\journalname{VLDB Journal}

\begin{document}

\title{MSAD: A Deep Dive into Model Selection for Time series Anomaly Detection}

\author{
	Emmanouil Sylligardos\textsuperscript{1} \and
	John Paparrizos\textsuperscript{2,3} \and
	Themis Palpanas\textsuperscript{4} \and
	Pierre Senellart\textsuperscript{1} \and
	Paul Boniol\textsuperscript{1}
}


\institute{
	Emmanouil Sylligardos \at
	\email{hiimsylli@gmail.com} \\
	\and
    \textsuperscript{1} DI ENS, ENS, PSL University, CNRS, Inria, Paris, France \\
    \textsuperscript{2} The Ohio State University, Columbus, OH, USA \\
    \textsuperscript{3} Aristotle University Thessaloniki, Thessaloniki, Greece \\
    \textsuperscript{4} Université Paris Cité, IUF Paris, France
}

\date{Received: 31 October 2024 / Accepted: 9 October 2025}

\maketitle

\begin{abstract}
	Anomaly detection is a fundamental task for time series analytics with important implications for the downstream performance of many applications. Despite increasing academic interest and the large number of methods proposed in the literature, recent benchmarks and evaluation studies demonstrated that no overall best anomaly detection methods exist when applied to very heterogeneous time series datasets. Therefore, the only scalable and viable solution to solve anomaly detection over very different time series collected from diverse domains is to propose a model selection method that will select, based on time series characteristics, the best anomaly detection methods to run. Existing AutoML solutions are, unfortunately, not directly applicable to time series anomaly detection, and no evaluation of time series-based approaches for model selection exists. Towards that direction, this paper studies the performance of time series classification methods used as model selection for anomaly detection. In total, we evaluate 234 model configurations derived from 16 base classifiers across more than 1980 time series, and we propose the first extensive experimental evaluation of time series classification as model selection for anomaly detection. Our results demonstrate that model selection methods outperform every single anomaly detection method while being in the same order of magnitude regarding execution time. This evaluation is the first step to demonstrate the accuracy and efficiency of time series classification algorithms for anomaly detection, and represents a strong baseline that can then be used to guide the model selection step in general AutoML pipelines. Preprint version of an article accepted at the VLDB Journal.
\end{abstract}

\ifdefempty{\vldbavailabilityurl}{}{
\vspace{.3cm}
\begingroup\small\noindent\raggedright\textbf{Artifact Availability:}\\
The source code, data, and/or other artifacts have been made available at \url{https://github.com/sylligardos/MSADv2}.
\endgroup
}

\section{Introduction}

\begin{figure}
    \centering
    \includegraphics[width=1\linewidth]{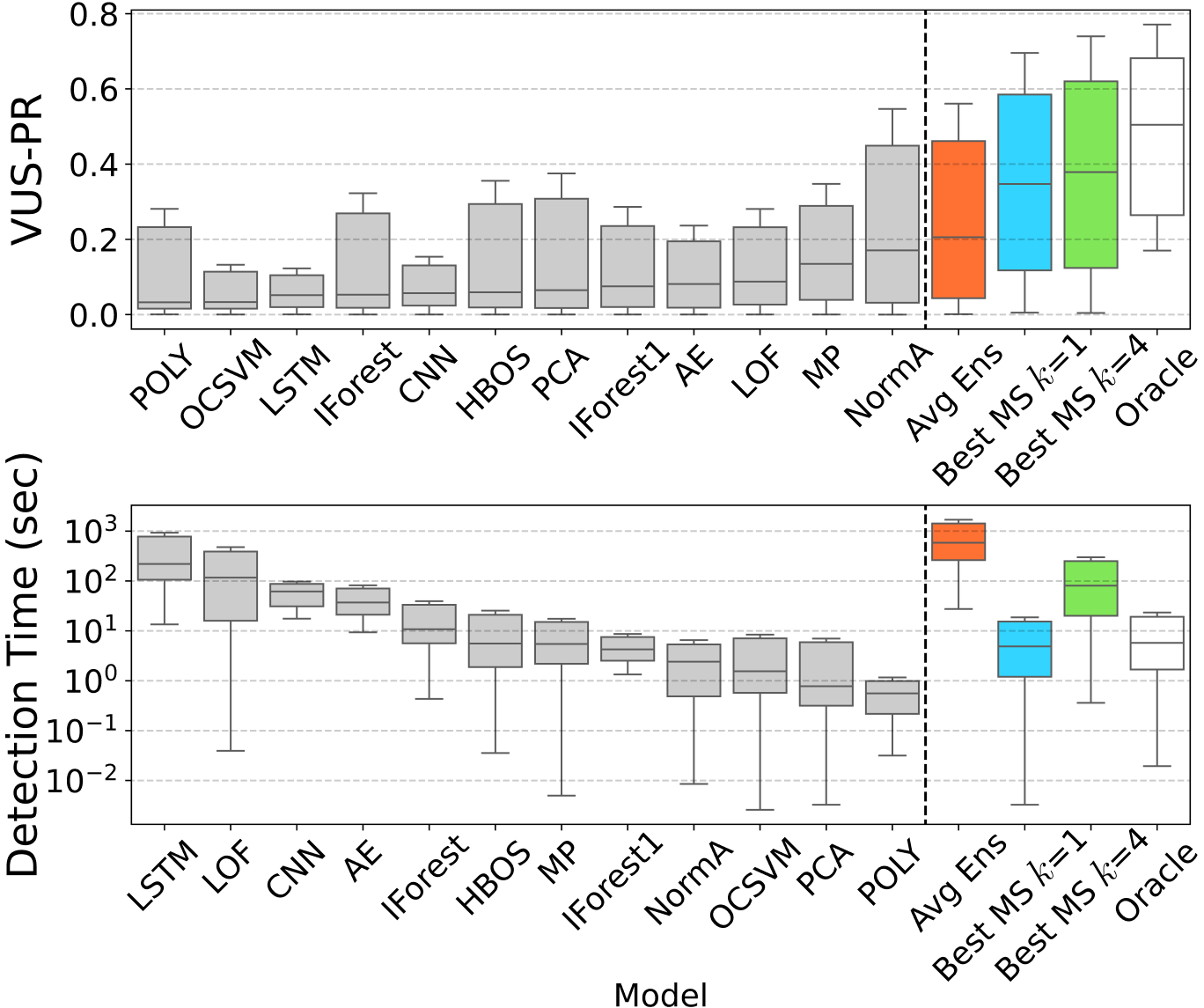}
    \caption{Summary of our evaluation on the TSB-UAD benchmark~\cite{10.14778/3529337.3529354} of model selection methods (best for $k=1$ in blue and $k=4$ in green) when compared to 12 anomaly detection methods and the Avg Ens (in orange).}
    \label{fig:intro_fig}
\end{figure}

Extensive collections of time-dependent measurements are a reality in every scientific domain~\cite{Palpanas2019,liu2023amir}. The recording of these measurements results in an ordered sequence of real-valued data points, commonly referred to as \textit{time series}~\cite{paparrizos2015k,bariya2021k,paparrizos2022fast}. Analyzing time series is becoming increasingly important in virtually every domain, including astronomy~\cite{huijse2014computational}, biology~\cite{bar2003continuous}, economics~\cite{lutkepohl2004applied}, energy sciences~\cite{bach2017flexible}, engineering~\cite{uehara2002extraction}, environmental sciences~\cite{goddard2003geospatial}, medicine~\cite{richman2000physiological}, neuroscience~\cite{biswal2010toward}, and social sciences~\cite{brockwell2016introduction}. Anomaly Detection (AD), in particular, has received ample academic and industrial attention~\cite{page1957problems,fox1972outliers}, and has become a significant problem that finds applications across a wide range of domains and situations. These applications share the same goal~\cite{statisticaloutliers,DBLP:conf/vldb/SubramaniamPPKG06,DBLP:conf/icdm/YehZUBDDSMK16}: analyzing time series to identify observations that do not correspond to an expected behavior inferred from previously observed data. In practice, anomalies can correspond to~\cite{aggarwal2017introduction}: (i) noise or erroneous data (e.g., broken sensors); or (ii) actual data of interest (e.g., abnormal behavior of the observed system). In both cases, detecting such types is crucial for many applications~\cite{IMSGroundtruth,DBLP:conf/healthcom/HadjemNK16}.

In recent years, many techniques have been proposed for Time Series Anomaly Detection (TSAD). Multiple surveys and benchmarks summarize and analyze the state-of-the-art proposed methods~\cite{blazquez2021review,10.14778/3529337.3529354,10.14778/3538598.3538602,10.14778/3551793.3551830,10.14778/3476249.3476307,wu2020current,7424283,jacob2020exathlon,Kim2021TowardsAR}. Such surveys and benchmarks provide a holistic view of anomaly detection methods and how they perform. Unfortunately, these benchmark and evaluation studies demonstrated that no overall best anomaly detection methods exist when applied to very heterogeneous time series (i.e., coming from very different domains). In practice, we observe that some methods outperform others on specific time series with either specific characteristics (e.g., stationary or non-stationary time series) or anomalies (e.g., point-based or sequence-based anomalies). 

To overcome the above limitation, ensembling solutions have been proposed~\cite{10.1145/2830544.2830549} that involve running all existing anomaly detection methods and averaging their anomaly scores. Figure~\ref{fig:intro_fig} shows that this solution (in orange) outperforms all individual existing techniques in the TSB-UAD benchmark (in grey)~\cite{10.14778/3529337.3529354,boniol2022theseus}. However, as shown in Figure~\ref{fig:intro_fig}, such solutions require running all methods, resulting in an excessive cost that is not feasible in practice. Additionally, blindly combining models without considering their performances may lead to reduced accuracy compared to a combination of only the top-performing detectors.

Therefore, the only scalable and viable solution for solving anomaly detection over very different time series collected from various domains is to propose a model selection method that selects, based on time series characteristics, the best anomaly detection methods to run. This topic has been tackled in several recent research works related to AutoML (Automated Machine Learning) for the general case of anomaly detection~\cite{NEURIPS2021_23c89427,ying2020automated} and also for time series~\cite{https://doi.org/10.48550/arxiv.2009.04395,https://doi.org/10.48550/arxiv.2210.01078}. Nevertheless, existing AutoML solutions require (i) a universal objective function among models, which is not applicable to anomaly detection methods; (ii) a predefined set of features, which is difficult to obtain for time series due to varying lengths and the lack of standardized featurization solutions; (iii) running multiple anomaly detection methods several times, which is prohibitively expensive in practice; or (iv) labeled anomalies, which (in contrast to classification tasks) are difficult to obtain. Therefore, more work is needed in order to render AutoML solutions applicable to TSAD. 

The objective is to train a classification model on time series for which we know in advance which anomaly detection method is the best. However, the lack of a benchmark with labeled time series has been a limiting factor for training robust model selection models (this only changed very recently~\cite{10.14778/3529337.3529354,10.14778/3538598.3538602,kdd21}). Therefore, there exists no experimental evaluation that measures the effectiveness of classification methods for the task of model selection for TSAD. Thus, such an evaluation is very important for determining which time series classification methods are accurate as model selection methods, and which solutions should be considered in \revtwo{out-of-distribution} settings (i.e., using model selection approaches on time series from domains that were not included in the training set). These results would help the design and effectiveness of general AutoML pipelines for time series.

Accordingly, in this paper, we evaluate the performance of time series classification methods used as model selectors for TSAD. \revone{Rather than relying on a single detector, we explore the benefits of combining multiple detectors per time series. Different anomaly detectors tend to specialize in certain data characteristics, as some detectors are better at capturing point anomalies, others at detecting sequence anomalies. For example, in Figure~\ref{fig:diversity} (a) we can see that NormA is undoubtedly better than CNN on ECG data, which behavior is reversed on the YAHOO dataset as shown in Figure~\ref{fig:diversity} (d). Overall, combining multiple detectors improves robustness, since real-world time series often contain more than a single type of anomalies.}

Thus, we build a framework~\footnote{A preliminary version of this work has appeared elsewhere~\cite{sylligardos2023choose}.} capable of proposing a single anomaly detector for a given time series, or a weighted combination of multiple anomaly detectors. In this framework we control the number of anomaly detection methods combined by incorporating a dynamic selection parameter $k$. By systematically adjusting the value of $k$, that is the number of detectors to combine, we study the trade-off between accuracy and runtime performance, demonstrating how a weighted subset of top-performing detectors enhances overall performance without the high costs associated with running all detectors. \revtwo{We evaluate the proposed pipeline in terms of accuracy and execution time in two distinct experimental settings: in-distribution and Out-Of-Distribution (OOD). In the in-distribution setting, the model is evaluated on time series from known domains, whereas in the OOD setting, the model is trained on multiple datasets spanning various domains and tested on a held-out dataset from a domain not seen during training (e.g., trained on electrocardiogram~\cite{Moody}, tested on robotics sensors measurements~\cite{5573462}). This OOD setting simulates a scenario where the inference data is entirely unseen, allowing us to assess the model's generalizability and performance in transfer learning contexts.}

Overall, we evaluate our framework on over 1980 time series and 12 anomaly detection methods from the recent anomaly detection benchmark TSB-UAD. To the best of our knowledge, our results demonstrate the first extended evaluation of time series classification methods as model selectors for TSAD. \revone{More specifically, we provide evidence that combining multiple anomaly detectors ($k > 1$) can significantly benefit the pipeline over choosing only a single detector ($k = 1$). Model selectors with $k > 1$ surpass (i) all stand-alone anomaly detection methods, (ii) the Averaging Ensemble (\textit{Avg Ens}), and (iii) all their single-detector counterparts with $k = 1$. The benefits \revone{are} accentuated in the \revtwo{OOD} setting where single-detector model selectors do not perform better than the naive Avg Ens, while multi-detector model selectors can surpass the Avg Ens performance while significantly reducing execution time.} Figure~\ref{fig:intro_fig} shows a summary of our experimental evaluation \revtwo{in the in-distribution setting, where the best model selection methods (shown in blue for $k = 1$ and green for $k=4$)} are up to 2.3$\times$ more accurate than the best anomaly detection method in the TSB-UAD benchmark and 1.9$\times$ more accurate than the ensembling solution mentioned above. In the \revtwo{OOD setting, combining detectors provides similar performance to the \textit{Avg Ens}, while significantly reducing execution time.} This framework provides the first step to demonstrate the accuracy and efficiency of time series classification algorithms for anomaly detection. It represents a strong baseline that can then be used to guide the choice of approaches for the model selection step in more general AutoML pipelines.
Overall, the paper is organized as follows:
\begin{itemize}
	\item We start with a detailed discussion of the relevant background and related work for anomaly detection in time series (Section~\ref{sec:background}).
    \item We cast the model selection problem for TSAD methods into a time series classification problem. We describe and study the need to evaluate time series classification methods for model selection (Section~\ref{sec:problem_def}). 
	\item We introduce our pipeline for model selection applied to anomaly detection in time series. As this pipeline is generic, we describe how it can be used with both feature-based classification methods, traditional time series classification methods, and deep learning-based methods (Section~\ref{sec:proposed}).
    \item As multiple detectors can be selected by our pipeline, we introduce two combination strategies. The first considers the prediction probabilities of the model selection methods, and the second uses a voting system across all the subsequences of the time series. The probabilities or the number of votes are then used to weight the detectors anomaly scores and generate the final anomaly score (Section~\ref{sec:proposed}).
	\item We describe our experimental framework (on top of TSB-UAD benchmark~\cite{10.14778/3529337.3529354}), and provide details on both anomaly detection methods and time series classification methods considered in this paper (Section~\ref{sec:exp}). We make all our material publicly available online~\cite{ourcode} and provide an interactive Web application~\cite{ourwebsite} for exploring our results. 
	\item We present an extensive experimental evaluation, measuring the anomaly detection accuracy and execution time (both training and inference) of model selection algorithms (Section~\ref{exp:overalleval}). We evaluate the influence of important parameters and the relationship between classification and anomaly detection accuracy (Sections~\ref{exp:windowlength}, \ref{exp:datasets}, and \ref{exp:detectionvsclass}). Moreover, we measure the transferability of model selection algorithms to new types of time series by testing multiple combinations of train and test datasets that do not contain the same kinds of time series (Section~\ref{exp:sup2unsup}).
    \item Finally, we conclude with the implications of our work and discuss possible future directions that could help improve both the accuracy and the execution time of our proposed pipeline (Section~\ref{sec:conclusions}).
\end{itemize}

\section{Background and Related Work}
\label{sec:background}

We first introduce formal notations (Section~\ref{sec:notation}), and then review in detail existing TSAD methods (Section~\ref{sec:ad_methods}) and discuss their limitations (Section~\ref{sec:limitation}).

\subsection{Time Series and Anomaly Score Notations}
\label{sec:notation}

\textbf{Time Series: }A time series $T \in \mathbb{R}^n $ is a sequence of real-valued numbers $T_i\in\mathbb{R}$ $[T_1,T_2,...,T_n]$, where $n=|T|$ is the length of $T$, and $T_i$ is the $i^{th}$ point of $T$. We are typically interested in local regions of the time series, known as subsequences. A subsequence $T_{i,\ell} \in \mathbb{R}^\ell$ of a time series $T$ is a continuous subset of the values of $T$ of length $\ell$ starting at position $i$, formally defined as $T_{i,\ell} = [T_i, T_{i+1},...,T_{i+\ell-1}]$. We then define a dataset $\mathcal{D}$, which is a set of time series. Note that the time series contained in $\mathcal{D}$ can be of diverse lengths. We define the size of $\mathcal{D}$ as $|\mathcal{D}|$.

\textbf{Anomaly Score Sequence: }For a time series $T \in \mathbb{R}^n $, an AD method (or detector) $D$ returns an anomaly score sequence $S_T$. For point-based approaches (i.e., methods that return a score for each point), we have $S_T \in \mathbb{R}^n$. For subsequence-based approaches (i.e., methods that return a score for each subsequence of a given length $\ell$), we have $S_T \in \mathbb{R}^{n-\ell}$ and $S_T = [{S_T}_1,{S_T}_2,...,{S_T}_{n-\ell}]$ with ${S_T}_i \in [0,1]$. In most applications, the anomaly score has to be the same length as the time series. Thus, for subsequence-based approaches, we define:

\begin{equation}
\small
    \begin{aligned}
        S_T =\, &[{S_T}_1]^{\ell/2}\, + [{S_T}_1, {S_T}_2, \ldots, {S_T}_{n-\ell}]\, + \\
                &[{S_T}_{n-\ell}]^{\ell/2} \text{ with } |S_T| = |T|
    \end{aligned}
\end{equation}
where $+$ denotes list concatenation, and $[x]^{\ell/2}$ represents a list containing the element $x$ repeated $\ell/2$ times.

\textbf{Anomaly Detection Accuracy: }For a time series \( T \in \mathbb{R}^n \), an AD method (or detector) \( D \) returns an anomaly score sequence \( D(T) = S_T \). The labels \( L \in [0,1]^n \) indicate with 0 or 1 if the points in \( T \) are normal or abnormal, respectively. We define \( Acc: \mathbb{R}^n \times \{0,1\}^n \rightarrow [0,1] \) as an accuracy function, for which \( Acc(D(T), L) \), namely the accuracy score, indicates how accurate \( D \) is (i.e., producing an score close to 1 when the label is 1, and close to 0 otherwise). The closer the accuracy score is to one, the better the detector.

\begin{figure}
    \centering
    \includegraphics[width=1\linewidth]{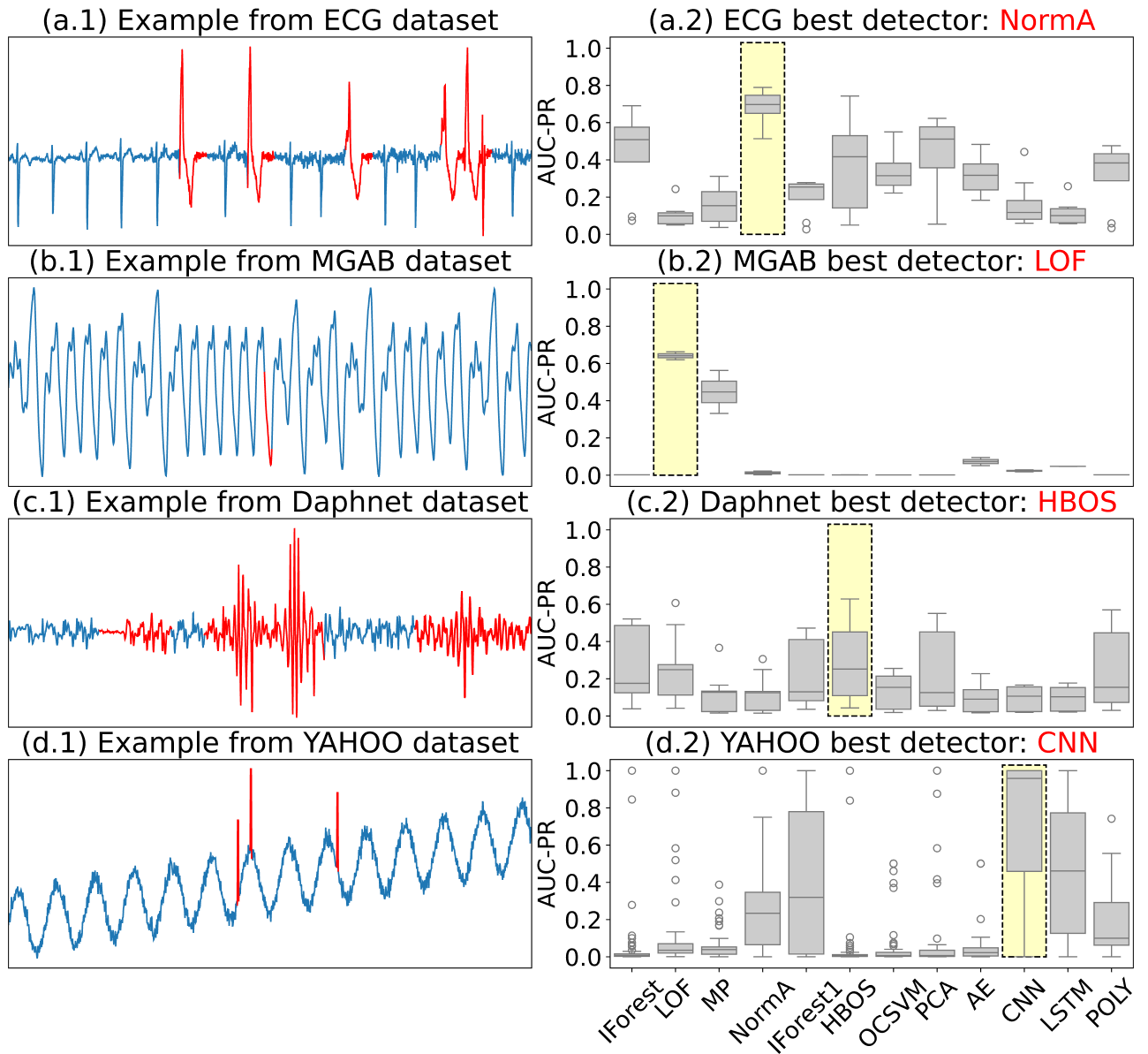}
    \caption{Accuracy of 12 detectors on 4 datasets.}
    \label{fig:diversity}
\end{figure}

\subsection{Anomaly Detection Methods for Time Series}
\label{sec:ad_methods}

Anomaly detection in time series is a crucial task for many relevant applications. Therefore, several methods (for diverse types of time series, or applications) have been proposed in the literature~\cite{boniol2024divetimeseriesanomalydetection}. One type of anomaly detection method is \textit{distance-based methods}, that analyze subsequences by utilizing distances to a given model to detect anomalies. In this category, we can identify three sub-categories. The first is \textit{discord-based}. These methods focus on the analysis of subsequences for the purpose of detecting anomalies in time series, mainly by utilizing nearest neighbor distances among subsequences \cite{DBLP:conf/icdm/YehZUBDDSMK16,DBLP:conf/edbt/Senin0WOGBCF15,Keogh2007,DBLP:journals/datamine/LinardiZPK20}. The second sub-category is \textit{proximity-based}. These methods focus on estimating the density of particular types of subsequences in order to either extract a normal behavior or isolate anomalies. Since a subsequence can be seen as a multidimensional point (with the number of dimensions corresponding to the subsequence length), general outlier detection methods can be applied for TSAD~\cite{Breunig:2000:LID:342009.335388,ma2020isolation}.The last category is \textit{clustering-based}, that contains methods using the distance to a given clustering partition to detect anomalies. In this sub-category, NormA, that first clusters data to obtain the normal behavior~\cite{norm,boniol_unsupervised_2021,boniol2021sand} have demonstrated strong performance.

While the previously mentioned methods compute their anomaly score on distances with raw time series element (such as subsequences), \textit{density-based methods} focus on detecting recurring or isolated behaviors by evaluating the density of the points or subsequences into a specific representation space. This category can be divided into four sub-categories, namely \textit{distribution-based}, \textit{graph-based}, \textit{tree-based}, and \textit{encoding-based}. Among them, Isolation Forest~\cite{liu_isolation_2008}, a \textit{tree-based} methods grouping points or subsequences into different trees, and Series2Graph, a \textit{graph-based} method that converts the time series into a graph to facilitate the detection of anomalies~\cite{Series2GraphPaper}, have been shown to work particularly well for TSAD task~\cite{Series2GraphPaper}. 

Furthermore, \textit{forecasting-based methods}, such as recurrent neural network-based ~\cite{malhotra_long_2015} or convolutional network-based~\cite{8581424}, have been proposed for this task. These methods use the past values as input, predict the following one, and use the forecasting error as an anomaly score. Such methods are usually trained on time series without anomalies, or make the assumption that the anomalies are significantly less frequent than the normal behaviors.

Finally, \textit{reconstruction-based methods}, such as autoencoder approaches~\cite{10.1145/2689746.2689747}, are trained to reconstruct the time series and use the reconstruction error as an anomaly score. As both forecasting and reconstruction-based categories detect anomalies using prediction errors (either forecasting or reconstruction error), we can group them into \textit{prediction-based methods}. 

\subsection{Limitations of Anomaly Detection Methods}
\label{sec:limitation}

Recently, several benchmarks and experimental evaluations for anomaly detection in time series have been proposed in the literature~\cite{10.14778/3538598.3538602,10.14778/3551793.3551830,kdd21}. Such benchmarks offer a comprehensive collection of time series from various domains and evaluate multiple methods within the categories mentioned above. However, these experimental evaluations led to the same conclusion: no method exists that outperforms all the others on all time series from various domains. Figure~\ref{fig:diversity}, which depicts the accuracy of 12 diverse AD methods\footnote{We use 12 methods that have been employed in previous studies ~\cite{10.14778/3529337.3529354,10.14778/3551793.3551830}. Note that other variations exist that may lead to improved results.} on four time series datasets, illustrates the conclusion above. In Figure~\ref{fig:diversity} (a.2), NormA is the most accurate model on the ECG dataset~\cite{Moody} (a time series example is depicted in Figure~\ref{fig:diversity} (a.1)). However, Local Outlier Factor (LOF)~\cite{Breunig:2000:LID:342009.335388}, and Matrix profile (MP)~\cite{DBLP:conf/icdm/YehZUBDDSMK16} are significantly outperforming NormA on the MGAB dataset~\cite{markus_thill_2020_3762385} (see Figure~\ref{fig:diversity} (b.2)), whereas CNN~\cite{8581424} is outperforming NormA, LOF, and MP on the YAHOO dataset~\cite{yahoo} (see Figure~\ref{fig:diversity} (d.2)). The following two reasons explain this difference in performance among datasets.

\subsubsection{\textbf{Heterogeneity in anomaly types}}

First, there are three types of time series anomalies: \textit{point}, \textit{contextual}, and \textit{collective} anomalies. \textit{Point} anomalies refer to data points that deviate remarkably from the rest of the data. Similarly, \textit{contextual} anomalies refer to data points within the expected range of the distribution (in contrast to point anomalies) but deviate from the expected data distribution, given a specific context (e.g., a window). For instance, Figure~\ref{fig:diversity} (d.1) illustrates a subsequence from the YAHOO dataset with \textit{contextual} \revone{anomalies. The values of the anomalies fall within the range of normal values, but are abnormal in the context of the distribution of the surrounding points. For these specific types of anomalies, \textit{reconstruction} and \textit{forecasting}-based methods are particularly accurate (as shown in Figure~\ref{fig:diversity} (d.2)).}

\textit{Collective} anomalies refer to sequences of points that do not repeat a typical (previously observed) pattern. The first two categories, namely, point and contextual anomalies, are referred to as \textit{point-based} anomalies, whereas \textit{collective} anomalies are referred to as \textit{subsequence} anomalies. For instance, Figure~\ref{fig:diversity} (a.1), (b.1), and (c.1) show three time series with sequence anomalies. However, even for time series belonging to the same anomaly type categories, we observe that the most accurate models are all different.  

\subsubsection{\textbf{Heterogeneity in time series structures}}

This diversity in model accuracy can be explained by other factors related to the time series structures. Indeed, on top of these categories mentioned above, the combination of them also matters. First, we need to differentiate time series containing \textit{single} anomalies from time series containing \textit{multiple} anomalies. Then, the \textit{multiple} time series category has to be divided into two subcategories, namely time series containing \textit{multiple different} and \textit{multiple similar} anomalies. For instance, methods based on neighbor distance computation such as LOF are very accurate in detecting \textit{single} or \textit{multiple different} anomalies, but less accurate for \textit{multiple similar}. To illustrate this point, Figure~\ref{fig:diversity} (a.2) depicts the results of 12 AD methods on the ECG dataset (that contains a large number of \textit{multiple similar} anomalies), for which LOF accuracy is low. On the contrary, Figure~\ref{fig:diversity} (b.2) depicts the results of the same 12 AD methods on the MGAB dataset (that contains \textit{multiple different} anomalies), for which LOF accuracy is high.

On top of the large variety of time series and anomaly characteristics mentioned above, time series can have distinct statistical characteristics, resulting in an even larger variability in the accuracy of AD methods. The latter can be the differences between \textit{stationary} (i.e., with a constant distribution of values over time) and \textit{non-stationary} (i.e., with a changing distribution of values over time) time series, or \textit{single normality} (i.e., time series containing only one normal behavior) and \textit{multiple normalities} (i.e., time series containing multiple normal behaviors) time series. 

\section{Motivation and Problem}
\label{sec:problem_def}

In this section, we describe solutions that can be applied to solve the limitations mentioned above, and we motivate the benefits of these solutions. Finally, we formally define the problem.

\subsection{Ensembling Detectors}

The first solution is to ensemble the anomaly scores produced by all the detectors. Multiple ensembling techniques have been proposed in the literature~\cite{10.1145/2830544.2830549} from which three main methods arise: (i) \textit{Averaging}: the average of the anomaly scores for each timestamp, (ii) \textit{Maximizing}: the maximum anomaly score for each timestamp (iii) \textit{Average of Maximum}: the average of the maximum for a randomly selected subset of detectors. \textit{Averaging} strategy is proven to be the more robust and low-risk strategy compared to the other two~\cite{10.1145/2830544.2830549}. Formally, the \textit{Averaging} strategy is defined as follows:

\revthree{\begin{definition}
    Given a time series $T$ of length $n$ and a set of detectors $\mathcal{B}$, the anomaly score sequence produced by the \textit{Averaging} strategy is defined as $AvgEns = [Avg_1,Avg_2, ..., Avg_n]$ where each element $Avg_i$ (for $i \in [1,n]$) is equal to $Avg_i= (1/|\mathcal{B}|)\sum_{D \in \mathcal{B}} D(T)_i$.
\end{definition}}

In the rest of the paper, we call the \textit{Averaging} strategy \textit{Averaging Ensemble (Avg Ens)}. As depicted in Figure~\ref{fig:intro_fig} (a), which shows the accuracy of detectors (in grey) and the \textit{Avg Ens} (in orange), we observe that such a strategy already outperforms all existing approaches. Nonetheless, such a method requires running all detectors to produce one ensembled anomaly score, resulting in a costly execution time (see Figure~\ref{fig:intro_fig} (b)). In a scenario with very long time series and an increasing number of detectors to consider, such an approach is not sustainable and feasible in practice.

\subsection{Model Selection}

A solution to tackle the limitations mentioned above is to apply model selection based on the characteristics of the time series. The primary objective is to train a model to automatically select the optimal combination of detectors (AD methods) for a given time series. In such a case, the user must run only $k$ models ($k$ ranging from 1 to the total number of available detectors), drastically reducing the execution time required (for $k$ significantly smaller than the total number of detectors). \revone{This allows users to benefit from the robustness of combining the complementary strengths of different detectors without having to run all of them as in \textit{Avg Ens}. The model selectors predict the most relevant detectors based on the time series characteristics and exclude the irrelevant ones.} 

This topic has been tackled in several recent papers related to AutoML (Automatic Machine Learning). Recent approaches, such as MetaOD~\cite{NEURIPS2021_23c89427,https://doi.org/10.48550/arxiv.2009.04395}, explored meta-learning to identify the best outlier detection algorithm on tabular datasets. These research works rely on the performance of pre-computed models on a subset of datasets to learn a mapping from the dataset's characteristics to the detectors' performance. Methods have been proposed to select models in an unsupervised way~\cite{https://doi.org/10.48550/arxiv.2210.01078}, but require running multiple models in advance, which (as \textit{Avg Ens}) limits the applicability due to high cost.

\subsection{Classification for Model Selection}

In general, for the specific case of time series, most of the work described above and future AutoML methods will rely on time series classification methods for the model selection step. In the simple case of model selection, where a single detector is predicted, the goal is to classify the time series into classes corresponding to the available AD methods. \revone{While this single-detector approach is sufficient for datasets with heterogeneous time series, it fails to capture the complementary strengths of multiple detectors like \textit{Avg Ens}. This limitation becomes particularly evident in complex time series with multiple patterns and anomaly types, where no single detector is consistently reliable.} Thus, to combine multiple detectors, the probability distribution inherently produced by the classification method is used as weights, and a weighted average combines the output of the detectors. However, no existing guidelines indicate which time series classification approach can be used for model selection. Thus, there is a need to evaluate and measure the benefit that time series classification approaches can bring to the anomaly detection task.

The first step is to evaluate the potential gain in accuracy that model selection could bring. To do this, recent TSAD benchmarks~\cite{10.14778/3529337.3529354,10.14778/3538598.3538602} can be used. We can evaluate the upper bound on the accuracy that model selection could reach on such benchmarks. Thus, we define a hypothetical model called $Oracle$, which, for a given time series, always selects the single most accurate anomaly detector to use. 

\revthree{Specifically, the aforementioned benchmarks provide, along with each time series, their ground-truth labels and the anomaly scores produced by multiple detectors. To compute the Oracle, we calculate the AUC-PR (Area Under the Precision-Recall curve) between the labels and each detector’s anomaly score for each time series. The Oracle then selects, for each time series, the detector that achieves the highest AUC-PR. This provides a theoretical upper bound for model selection performance, simulating the perfect model selector.} We are not creating a theoretical upper bound that combines more than one detector, as the $Oracle$ already serves the purpose sufficiently (see Figure~\ref{fig:intro_fig} (a)). Nonetheless, we hope that in the future, this upper bound will be surpassed and the need for another upper bound will arise. Formally, $Oracle$ is defined as follows:
\begin{definition}
    Given a dataset $\mathcal{D}$ composed of time series $T$ and labels $L$ (with the length of the time series $|T|=n$ non-constant for all time series in $\mathcal{D}$), and a set of detectors $\mathcal{B} = \{D_1, D_i, ..., D_m\}$ with the number of detectors defined as $|\mathcal{B}|=m$, $Oracle(T)= \operatorname*{argmax}_{D \in \mathcal{B}} \big\{Acc\big(D(T),L\big)\big\}$ 
\end{definition}

We refer to $Oracle$, the hypothetical model $Oracle(T)$, as applied to all $T$ in a given benchmark. For example, Figure~\ref{fig:intro_fig} shows in white the accuracy of $Oracle$ applied to the TSB-UAD benchmark~\cite{10.14778/3529337.3529354} and demonstrates that a perfect model selection method outperforms the best detector in TSB-UAD and the \textit{Avg Ens} by a factor of 3 and 2.5 accordingly. This large improvement in accuracy and execution time confirms the potential benefits of model selection applied to TSAD. Thus, there is a need to evaluate the performance of existing time series classification methods when used as model selection algorithms and how close such methods can get to the $Oracle$.

\subsection{Problem Formulation}

Therefore, based on the limitations and the motivation listed above, we can formalize the problem of model selection as follows:

\begin{problem}
    \label{prob:probdef}
    Given a dataset $\mathcal{D}$ composed of time series $T$ (with the length of the time series $|T|=n$ non-constant for all time series in $\mathcal{D}$) and a set of detectors $\mathcal{B} = \{D_1, D_2, ..., D_m\}$ with the number of detectors defined as $|\mathcal{B}|=m$. We want to build a model selection method $\mathcal{M}$ that takes a time series $T \in \mathcal{D}$ and returns a set of weights $W_T = \{w_1, w_2, ..., w_m\}$ (formally $\mathcal{M}: \mathbb{R}^n \rightarrow \mathbb{R}^m$) such that the anomaly score $S_T$ is a weighted combination of the scores from the detectors in $\mathcal{B}$:
    \begin{align*}
    \small
        S_T = \sum_{i=1}^{m} w_i D_i(T), \text{ where } \sum_{i=1}^{m} w_i = 1 \text{ and } w_i \geq 0
    \end{align*}
     For a given parameter $k$, which specifies the number of detectors to combine, $\mathcal{M}$ will ensure that only $k$ out of the $m$ weights are non-zero. The goal is to maximize the accuracy of the combined anomaly score $S_T$ with respect to the label $L$:
    \begin{align*}
    \small
        \mathcal{M}(T) = \operatorname*{argmax}_{W_T} \bigg\{Acc\bigg(\sum_{i=1}^{m} w_i D_i(T), L\bigg)\bigg\}
    \end{align*}
\end{problem}
In practice, we do not have the label $L$. Therefore, the objective is to build a model $\mathcal{M}$ that estimates the equation above.
Moreover, as the input of $\mathcal{M}$ is a time series and the output is a set of weights for the detectors in $\mathcal{B}$, the problem can be seen as a time series classification problem for which the classes are the detectors in $\mathcal{B}$ and the set of weights is the produced probability distribution over the classes. Thus, the only requirement is to have computed all $Acc(D(T),L)$ for all $T \in \mathcal{D}$ and all $D \in \mathcal{B}$ and use it as a training set.

\subsection{Objectives}
\label{sec:objective}

In summary, our goal is to answer the following questions:
\begin{itemize}
	\item \textbf{Classification as Model selection}: How do classification methods compare to individual detectors and the $Oracle$?
    \revone{\item \textbf{Single vs. multiple detectors}: Is combining multiple detectors (i.e., $k > 1$) better than selecting a single best one?}
	\item \textbf{Ensembling or selecting}: Is selecting $k$ detectors automatically more accurate than ensembling them? How large $k$ should be to outperform ensembling?
	\item \textbf{Features or Raw values}: Should we use time series features or the raw time series values to predict which detectors to use?
	\item \textbf{Out-Of-Distribution}: What happens when the model selection approach is trained on some datasets and tested on entirely new ones? Are the answers from the previous questions still valid? 
\end{itemize}

\noindent We now describe our pipeline and experimental evaluation to answer the questions listed above. 

\section{MSAD: Proposed Pipeline}
\label{sec:proposed}

\begin{figure*}
    \centering
    \includegraphics[width=0.8\linewidth]{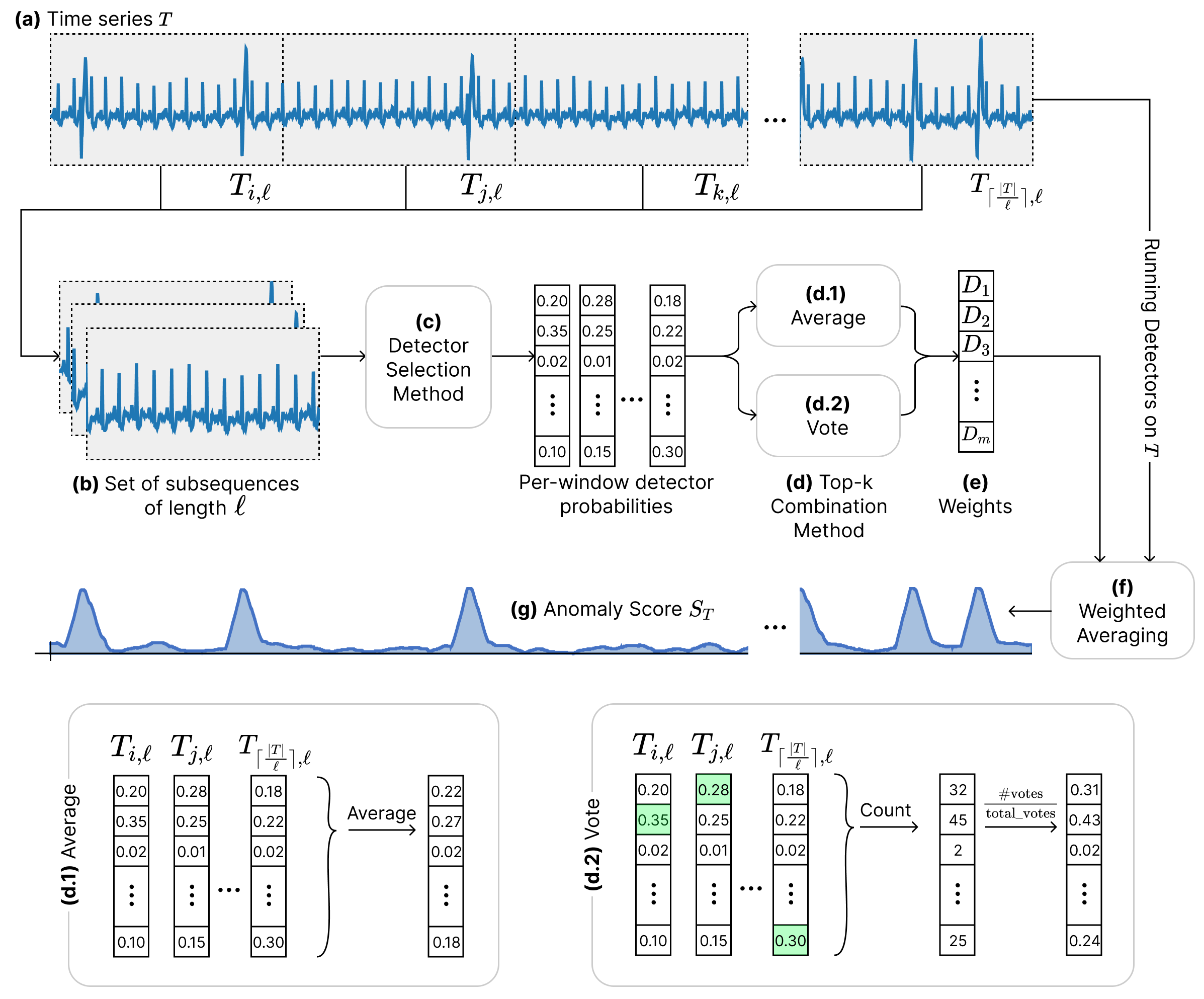}
    \caption{\revthree{Proposed pipeline for the method selection}}
    \label{fig:proposed_work}
\end{figure*}

In the following section, we provide a comprehensive explanation of the proposed pipeline. This pipeline involves a sequence of preprocessing and postprocessing steps to ensure that the inputs to the model selection algorithms are of equal length. The proposed pipeline, illustrated in Figure~\ref{fig:proposed_work}, consists of the following steps: (i) \textbf{Preprocessing step}: Extraction of subsequences of the same length (Figure~\ref{fig:proposed_work} (b)), (ii) \textbf{Prediction step}: Producing the probability distribution over the available classes for each subsequence (Figure~\ref{fig:proposed_work} (c)), and (iii) \textbf{Combination step}: Transforming the probability distributions into a sequence of weights (Figure~\ref{fig:proposed_work} (d)), one for each class, and combining the scores from individual detectors according to these weights (Figure~\ref{fig:proposed_work} (f)). In the following section, we describe the three steps mentioned above in detail.

\begin{table*}[tb]
    \centering
    \caption{Summary of datasets, methods, and measures used in this experimental evaluation.}
        \scalebox{0.82}{
        \begin{tabular}{|c|c|}
        \hline
        \rowcolor{Gray}
        \textbf{Datasets} & \textbf{Description} \\
        \hline
        Dodgers~\cite{10.1145/1150402.1150428} & unusual traffic after a Dodgers game \textbf{(1 time series)}\\ 
        \hline
        ECG~\cite{Moody} & standard electrocardiogram dataset \textbf{(52 time series)}\\ 
        \hline
        IOPS~\cite{IOPS} & performance indicators of a machine \textbf{(58 time series)}\\ 
        \hline
        KDD21~\cite{kdd21} & composite dataset released in a recent SIGKDD 2021 \textbf{(250 time series)}\\ 
        \hline
        MGAB~\cite{markus_thill_2020_3762385} & Mackey-Glass time series with non-trivial anomalies \textbf{(10 time series)}\\ 
        \hline
        NAB~\cite{ahmad_unsupervised_2017} &  Web-related real-world and artificial time series \textbf{(58 time series)}\\ 
        \hline
        SensorScope~\cite{YAO20101059} & environmental data \textbf{(23 time series)}\\ 
        \hline
        YAHOO~\cite{yahoo} & \makecell{ time series based on Yahoo production systems \textbf{(367 time series)}}\\ 
        \hline
        Daphnet~\cite{5325884} & acceleration sensors on Parkinson's disease patients \textbf{(45 time series)}\\ 
        \hline
        GHL~\cite{filonov2016multivariate} & Gasoil Heating Loop telemetry \textbf{(126 time series)}\\ 
        \hline
        Genesis~\cite{vonBirgelen2018} & portable pick-and-place demonstrator \textbf{(6 time series)}\\ 
        \hline
        MITDB~\cite{Moody} & ambulatory ECG recordings \textbf{(32 time series)}\\ 
        \hline
        OPPORTUNITY~\cite{5573462} & motion sensors for human activity recognition \textbf{(465 time series)}\\ 
        \hline
        Occupancy~\cite{CANDANEDO201628} & temperature, humidity, light, and CO2 of a room \textbf{(10 time series)}\\ 
        \hline
        SMD~\cite{10.1145/3292500.3330672} & Server Machine telemetry \textbf{(281 time series)}\\ 
        \hline
        SVDB~\cite{greenwald_improved_1990} & ECG recordings \textbf{(115 time series)}\\ 
        \hline
        \hline
        \rowcolor{Gray}
        \textbf{Anomaly Detection} & \textbf{Description} \\
        \hline
        IForest~\cite{liu_isolation_2008} & \makecell{constructs trees based on random splits. The nodes (i.e., subsequences) with shorter depth are labeled as anomalies}\\
        \hline
        IForest1~\cite{liu_isolation_2008}  & same as IForest, but each point (individually) is used as input\\
        \hline
        LOF~\cite{Breunig:2000:LID:342009.335388} & computes the ratio of the neighboring density to the local density\\ 
        \hline
        MP~\cite{yeh_time_2018} & \makecell{detects abnormal subsequences with the largest nearest neighbor distance} \\ 
        \hline
        NormA~\cite{boniol_unsupervised_2021} & \makecell{identifies normal patterns using clustering and computes weighted distance to the normal patterns} \\ 
        \hline
        PCA~\cite{aggarwal_outlier_2017} & \makecell{projects data to a lower-dimensional hyperplane and computes distance between subsequences and this plane} \\ 
        \hline
        AE~\cite{10.1145/2689746.2689747} & \makecell{trained to encode and reconstruct the data, and outliers are expected to have larger reconstruction errors} \\ 
        \hline
        LSTM-AD~\cite{malhotra_long_2015} & \makecell{use an LSTM network to forecast the following value. The error prediction is then used to identify anomalies}\\ 
        \hline
        POLY~\cite{li_unifying_2007} & \makecell{fits a polynomial to forecast time series values. Outliers are detected with prediction error} \\ 
        \hline
        CNN~\cite{8581424} & \makecell{forecasts the time series values with a convolutional neural network. The anomaly score is the prediction error} \\ 
        \hline
        OCSVM~\cite{scholkopf_support_1999} & \makecell{is a support vector method that fits the normal training dataset and finds the normal data's boundary}\\ 
        \hline
        HBOS~\cite{goldstein2012histogram} & \makecell{builds a histogram for the time series. The anomaly score is the  inverse of the height of the bin} \\
        \hline
        \hline
        \rowcolor{Gray}
        \textbf{Model Selection} & \textbf{Description} \\
        \hline
        SVC~\cite{10.1145/130385.130401} \trianglemarker{featurebased} &  \makecell{maps instances to points in space to maximize the gap between classes}\\
        \hline
        Bayes~\cite{Zhang2004TheOO} \trianglemarker{featurebased} & \makecell{uses Bayes' theorem to classify a point using each class posterior probabilities} \\
        \hline
        MLP~\cite{Hinton1989ConnectionistLP} \trianglemarker{featurebased} & \makecell{consists of multiple layers of interconnected neurons} \\
        \hline
        QDA~\cite{Geisser1964PosteriorOF} \trianglemarker{featurebased} & is a discriminant analysis algorithm for classification problems \\
        \hline
        AdaBoost~\cite{10.5555/646943.712093} \trianglemarker{featurebased} & \makecell{is a meta-algorithm using boosting technique with weak classifiers} \\
        \hline
        Decision Tree~\cite{Hunt1966ExperimentsII} \trianglemarker{featurebased} & \makecell{is an approach that splits data points into separate leaves based on features} \\
        \hline
        Random Forest~\cite{598994} \trianglemarker{featurebased} & \makecell{is a set of Decision Trees fed with random samples and features} \\
        \hline
        kNN~\cite{Fix1989DiscriminatoryA} \trianglemarker{featurebased} & \makecell{assigns the most common class among its k nearest neighbors} \\
        \hline
        Rocket~\cite{dempster2021minirocket} \xmarker{rocket} & \makecell{transforms time series using a set of convolutional kernels, creating features used to train a linear classifier} \\
        \hline
        ConvNet~\cite{DBLP:journals/corr/WangYO16} \circlemarker{convbased} & \makecell{uses convolutional layers to learn spatial features from the input data} \\
        \hline
        ResNet~\cite{DBLP:journals/corr/WangYO16} \circlemarker{convbased} & \makecell{is a ConvNet with residual connections between convolutional block} \\
        \hline
        Inception Time~\cite{fawaz2020inceptiontime} \circlemarker{convbased} & \makecell{is a combination of ResNets with kernels of multiple sizes} \\
        \hline
        SiT-conv~\cite{dosovitskiy2020image} \rectanglemarker{transbased} & \makecell{is a transformer architecture with a convolutional layer as input} \\
        \hline
        SiT-linear~\cite{dosovitskiy2020image} \rectanglemarker{transbased} & \makecell{is a transformer architecture for which non-overlapping subsequences are linearly projected into the embedding space} \\
        \hline
        SiT-stem~\cite{xiao2021early} \rectanglemarker{transbased} & \makecell{is a transformer architecture with convolutional layers with increasing dimensionality as input} \\
        \hline
        SiT-stem-ReLU~\cite{wang2022scaled} \rectanglemarker{transbased} & \makecell{is similar to SiT-stem but with Scaled ReLU} \\
        \hline
        \hline
        \rowcolor{Gray}
        \textbf{Evaluation} & \textbf{Description} \\
        \hline
        Classification Accuracy &  \makecell{the number of correctly selected methods divided by the total number of time series}\\
        \hline
        AUC-PR~\cite{10.1145/1143844.1143874} &  Area under the Precision-Recall curve\\
        \hline
        VUS-PR~\cite{10.14778/3551793.3551830} &  \makecell{Volume under the Precision-Recall surface (obtained from different length of a buffer region surrounding the anomalies)}\\
        \hline
        Training Time & \makecell{number of seconds required to train a model selection method} \\
        \hline
        Selection Time & \makecell{number of seconds required to predict the best model to use} \\
        \hline
        Detection Time & \makecell{number of seconds required to compute an anomaly score (i.e., selection time plus detector execution)} \\
        \hline
        \end{tabular}
        } 
        \label{SymbolTable}
\end{table*}

\subsection{Preprocessing Step}
\label{sec:preprocessing}
Time series classification can be performed with three different strategies: (i) treating the entire time series as one sample, (ii) dividing the time series into overlapping subsequences, (iii) dividing the time series into shifting subsequences (i.e., non-overlapping subsequences). The first strategy is straightforward, as each time series is treated as a single observation. Nevertheless, not all classifiers can handle variable-length inputs, and training such models can be computationally intensive (i.e., batches of time series cannot be treated in parallel). The second strategy involves dividing the time series into overlapping subsequences (of a given window length $\ell$). Despite possible loss of information, it forces each input of the methods to be the same length ($\ell$), allowing simpler and faster computation when performed in parallel. In the third strategy, we divide time series into non-overlapping subsequences (of a given length $\ell$), removing redundant information in overlapping subsequences. The latter might lead to separate anomalies into multiple windows, but significantly reduces the number of inputs generated by the second strategy and significantly accelerates the training and inference time. For these reasons, we chose the third strategy.

\revthree{Thus, the time series of length $|T|$ are divided into $\mathbb{T}_l$ non-overlapping subsequences of length $\ell$. When the length of the time series is not divided evenly by the window length $\ell$, the remainder is added with an overlap between the first two windows. Formally:
\begin{equation*}
\small
    \mathbb{T}_\ell =
        \begin{cases}
            \left\{ T_{i \cdot \ell,\; \ell} \;\middle|\; i \in \left[0, n \right] \right\}, \text{ if } |T| \bmod \ell = 0 \\
            \\
            \left\{ T_{0,\; \ell} \right\} \cup \left\{ T_{|T| - n \cdot \ell + i \cdot \ell,\; \ell} \;\middle|\; i \in \left[0,\; n - 1 \right] \right\}, \text{otherwise}
        \end{cases}
\end{equation*}
}

With $n = \left\lceil \frac{|T|}{\ell} \right\rceil - 1$. We expect the length $\ell$ to have an impact on the anomaly detection accuracy. We thus test multiple length values and measure their influence (on accuracy and execution time) in Section~\ref{sec:exp}.

At this point, we preprocessed the time series into subsequences of equal length. We now discuss the label (i.e., the best detector to apply) attribution. For that matter, we use the TSB-UAD benchmark~\cite{10.14778/3529337.3529354} that contains 12 different AD methods. We compute the 12 methods for each time series and attribute the most accurate (based on AUC-PR) detector as the label. Then, the produced subsequences share the same label as the time series they originate from. \revthree{This design choice aims to learn global signal characteristics, rather than localized anomaly behavior, which we found to result in better classifier performance. We experimented with assigning different labels per subsequence, but this led to significantly worse results and was also challenging to evaluate reliably.} This labeled dataset can then be used to train classification methods and divided into the train, test, and validation sets. It is important to note that although each time series produces multiple samples (i.e., subsequences), these samples should not be mixed between train, validation, and test sets. Indeed, too strong similarities between subsequences that belong to the same time series, if contained in both the train, validation, and the test, can lead the classification model to overfit or create an illusion of accuracy. Therefore, we guarantee that the intersection between the train, validation, and test sets, regarding which time series the corresponding subsequences originate from, is empty.

\subsection{Time Series Classification Approaches}

\begin{figure}
    \centering
    \includegraphics[width=0.9\linewidth]{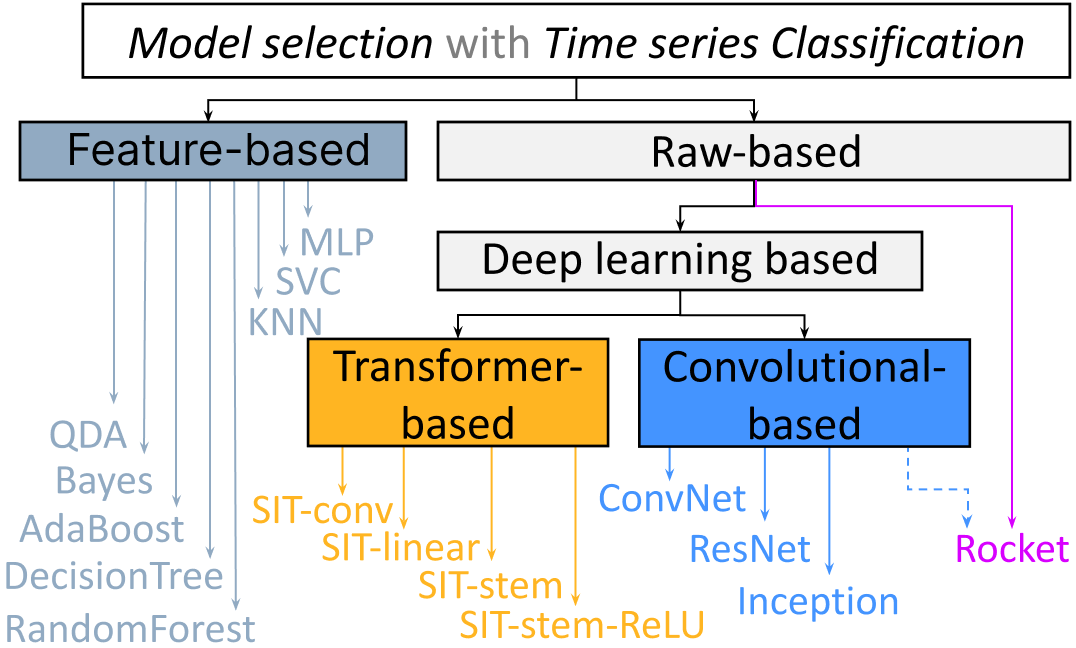}
    \caption{Taxonomy of time series classification approaches used as model selection methods. We use the same color code for each class in all figures in the paper.}
    \label{fig:taxonomy}
\end{figure}

In this section, we describe the time series classifier approaches that we use as model selection methods. As many approaches have been proposed in the literature, we restrict our experimental evaluation to two main categories: (i) feature-based and (ii) raw-based methods. In addition, the second category can be divided into two sub-categories: (i) convolutional-based and (ii) transformer-based. It is worth noting that raw-based methods also utilize features for classification, as the extraction of these features is performed automatically within the network. Despite this, we classify them as raw-based due to the nature of their input. Figure~\ref{fig:taxonomy} illustrates a simplified taxonomy of the methods considered, and we describe them in the following section.

\subsubsection{Feature-based classification}
The main idea regarding feature-based classification is to use the dataset of time series (or subsequences of time series) to create a dataset whose samples are described by features common to all samples. Using the feature-based dataset, we then employ traditional machine learning classifiers to classify each time series. We use the TSFresh~\cite{CHRIST201872} (Time Series Feature extraction based on scalable hypothesis tests) to extract each subsequence's features. The latter is used for automated time series feature extraction and selection based on the FRESH algorithm~\cite{christ2016distributed}. More specifically, it automatically selects relevant features for a specific task. This is achieved using statistical tests, time series heuristics, and machine learning algorithms. The TSFresh package provides three options for automated feature extraction, namely, (i) \textit{ comprehensive}, (ii) \textit{ efficient}, and (iii) \textit{ minimal}. The first two options provide \revone{700 features} and the latter provides only 9. For scalability reasons (the datasets can reach millions of subsequences), we consider the \textit{minimal} option in this paper.

Moreover, the objective is not to evaluate Feature-based classifiers \textit{per se}, but rather to evaluate the ability of TSFresh to extract meaningful features for time series classification (and model selection for anomaly detection, in particular). In this paper, we consider the following classification approaches.

\noindent\textbf{[SVC]}
A Support Vector Classifier (SVC)~\cite{10.1145/130385.130401} is a classifier that maps instances in space in order to maximize the width of the gap between the classes. New instances are mapped into the same space and classified according to which side of the gap they fall. 

\noindent\textbf{[Bayes]}
The naive Bayes classifier~\cite{Zhang2004TheOO} uses Bayes' theorem to predict the class of a new instance based on prior probabilities and class-conditional probabilities. The prediction is made by computing the posterior probabilities for each class.

\noindent\textbf{[MLP]}
A Multi Layer Perceptron (MLP)~\cite{Hinton1989ConnectionistLP} is a fully connected neural network.

\noindent\textbf{[QDA]}
A Quadratic Discriminant Analysis (QDA)~\cite{Geisser1964PosteriorOF} Classifier is a linear discriminant analysis algorithm. The prediction is made by computing the discriminant functions for each class.

\noindent\textbf{[AdaBoost]}
AdaBoost~\cite{10.5555/646943.712093} is a boosting ensemble machine learning algorithm for solving classification problems. It creates a sequence of weak classifiers, where each classifier is trained on a weighted sample of the dataset. The prediction is made by combining the predictions of all classifiers, weighted by their accuracy.

\noindent\textbf{[Decision Tree]}
A Decision Tree Classifier~\cite{Hunt1966ExperimentsII} is a tree-based method that represents a sequence of decisions based on the features of the dataset. To classify a new instance, the algorithm follows the decisions in the tree to reach a leaf node associated with a class.

\noindent\textbf{[Random Forest]}
A Random Forest~\cite{598994} is an ensemble machine learning algorithm that combines multiple decision trees, where each tree is built using a random subset of the features and a random sample of the data. The final class prediction for a new instance results from the aggregation of the predictions of all trees.

\noindent\textbf{[kNN]}
A kNN classifier~\cite{Fix1989DiscriminatoryA} is a method that classifies instances based on their distance to other instances in a training set. The algorithm assigns the new instances to the class with the most number of closest neighbors among the $K$ nearest data points. 

\subsubsection{Raw-based classification}

Instead of using extracted features to perform classification, the raw values of the time series can be used directly. While features are efficient for homogenizing time series datasets (e.g., setting a constant number of features for variable length time series), this approach might hide important information in the shape of consecutive values. Consequently, many approaches that use raw-values time series have been proposed. \revthree{However, it should be noted that, although raw-based methods use the raw time series as input, they still perform feature extraction internally. Features are usually extracted within the first steps of the model and it is a learned process, in contrast to the static feature extraction used in feature-based methods.
While other relevant classification methods could also be considered in this category~\cite{DBLP:conf/icdm/YehZUBDDSMK16}, we have carefully selected those that have demonstrated strong performance in recent evaluations~\cite{10.1007/s10618-024-01022-1}. Our choices also aim to cover the broadest possible range of methods, ensuring diversity across different approaches.}

\noindent\textbf{[Rocket]}
Among the recent raw-values methods, Mini\-Rocket~\cite{dempster2021minirocket} is one of the state-of-the-art time series classification methods. The latter consists of a feature extraction step and a classification step. More specifically, MiniRocket works by transforming input time series using a small, fixed set of convolutional kernels and using the transformed features to train a logistic regression classifier (using stochastic gradient descent). We refer to MiniRocket as Rocket.

\subsubsection{Convolutional-based classification}

Convolutional-based approaches take as input raw-values of time series and have been shown to be accurate for time series classification~\cite{DBLP:conf/sigmod/BoniolMRP22}.

\noindent\textbf{[ConvNet]}
A Convolutional Neural Network (CNN) \cite{DBLP:journals/corr/OSheaN15} is a type of deep learning neural network widely used in image recognition that is specially designed to extract patterns through data with a grid-like structure, such as images, or time series. A CNN uses convolution, where a filter is applied to a sliding window over the time series. The ConvNet architecture proposed in~\cite{DBLP:journals/corr/WangYO16} is composed of three stacked Convolutional blocks followed by Global Average Pooling (GAP), and a Softmax activation function. Each Convolutional block is composed of a convolutional layer (used with a kernel length of $3$) followed by a batch normalization layer, followed by a ReLU activation function is applied.

\noindent\textbf{[ResNet]}
The Residual Network (ResNet) architecture~\cite{https://doi.org/10.48550/arxiv.1512.03385} was introduced to address the gradient vanishing problem encountered in large CNNs~\cite{simonyan2015a}. A ResNet is composed of several blocks connected together with residual connections (i.e., identity mapping). For time series classification, a ResNet architecture has been proposed in~\cite{DBLP:journals/corr/WangYO16}, and has demonstrated strong classification accuracy~\cite{fawaz2019deep}. It is the same architecture as the previously described ConvNet, with additional residual connections between convolutional blocks.

\noindent\textbf{[InceptionTime]}
The model consists of a network using residual connections and convolutional layers with kernels of variable lengths~\cite{fawaz2020inceptiontime}. Such a network uses three Inception blocks that replace the traditional residual blocks that we can find in a ResNet architecture. Each Inception block consists of a concatenation of convolutional layers using different sizes of filters. For each block, the time series is fed to three different 1D convolutional layers with different kernel sizes (10, 20, and 40) and one Max-Pooling layer with kernel size 3. The last step consists of concatenating the previous four layers along the channel dimension and applying a ReLU activation function to the output, followed by batch normalization. The convolutional layers have 32 filters and a stride parameter of 1.

\subsubsection{Transformer-based classification}

Transformer-based approaches were initially introduced for Natural Language Processing~\cite{vaswani2017attention}. Such methods can easily be adapted for time series classification tasks, and in this paper we propose SiT (Signal Transformer), an extension of a recent computer vision transformer approach~\cite{dosovitskiy2020image}. SiT first starts by projecting the input to the latent space with an embedding step. After the embedding step, the input is mapped to a $D$ dimensional space (we use $D=256$ in the rest of the paper) that serves as input to an encoder. For SiT, we use an encoder originally proposed for computer vision tasks~\cite{vaswani2017attention} that consists of multiple blocks. Each block has an alternating multi-headed self-attention block and a feed-forward layer, both preceded by a normalization step and a residual connection. We now describe the different embedding steps in detail in the following paragraphs. In the experimental evaluation, we consider the SiT architecture with the four embeddings as four different methods.

\noindent\textbf{[SiT-conv]}
This embedding uses a single convolutional layer to map the time series into the latent space. The convolutional layer has a kernel and stride of the same length (we use a length of 16 throughout the rest of the paper), essentially taking non-overlapping steps over the time series. Finally, the convolutional layer has $D$ filters to match the input dimension of the SiT encoder.

\noindent\textbf{[SiT-linear]}
The linear embedding~\cite{dosovitskiy2020image} splits the input time series into non-overlapping subsequences of length $l_{SiT}$ (we use $l_{SiT}=16$ in the rest of the paper). Then, each patch is linearly projected into $D$ dimensions to match the input dimension of the SiT encoder.

\noindent\textbf{[SiT-stem]}
The stem embedding~\cite{xiao2021early} consists of 3 convolutional layers with a kernel length of 3, a stride length of 2, and a number of filters equal to 3, 5, and 7, respectively. These three convolutional layers are then followed by a last convolutional layer with $D$ dimensions and a kernel and stride length equal to 1. This embedding was initially proposed to overcome unstable behavior while training because of its early visual processing step. 

\noindent\textbf{[SiT-stem-ReLU]}
Similarly to the previous embedding, the stem-ReLU embedding~\cite{wang2022scaled} consists of 4 convolutional layers with kernel lengths of 7, 3, 3, 8, stride lengths of 2, 1, 1, 8, and padding of 3, 1, 1, 0. The number of filters for each convolutional layer is 3, except the last one with $D$ filters to match the SiT encoder's dimension.

\subsection{Combining Anomaly Scores}
\label{sec:combining}

\revone{Unlike selecting a single detector, which limits anomaly detection to a single detector's perspective, combining detectors allows us to leverage complementary strengths across multiple subsequences and anomaly types. This combination process begins by using the model selector to generate probability distributions over the detectors for each subsequence of a time series (as shown in Figure~\ref{fig:proposed_work} (c)).} These distributions reflect the model's prediction of each detector’s relevance to each subsequence. To combine these distributions into a final set of weights for the entire time series, we employ two methods:

\textbf{Average Strategy} (Figure~\ref{fig:proposed_work} (d.1)): We compute the average probability distribution over all subsequences. This results in a single average probability distribution that reflects the overall likelihood of each detector being the best choice. Formally, the aggregated probability of a given detector $D_j$, noted $\bar{P}_{D_j}$, using the \textit{Average} strategy is defined as follows:

\begin{equation}
\small
    \bar{P}_{D_j} = \frac{1}{n} \sum_{i=1}^{n} P^{(i)}_{D_j}, \text{where } n = \big\lceil \frac{|T|}{\ell} \big\rceil
    \label{eqmean}
\end{equation}
The final weights are computed by selecting the top-$k$ detectors and re-normalizing their probabilities.

\textbf{Vote Strategy} (Figure~\ref{fig:proposed_work} (d.2)): For each subsequence, we extract the detector with the highest probability, effectively casting a vote for that detector. We aggregate these votes across all subsequences. Formally, the aggregated vote for a given detector $D_j$, noted $\bar{V}_{D_j}$, using the \textit{Voting} strategy is defined as follows:

\begin{equation}
\small
    \bar{V}_{D_j} = \sum_{i=1}^{n} \mathds{1}_{\big[P^{(i)}_{D_j} = max_{k \in [1,m]} P^{(i)}_{D_k} \big]},  \text{where } n = \big\lceil \frac{|T|}{\ell} \big\rceil
    \label{eqvote}
\end{equation}
As in the average method, we compute the final weights by selecting the top-$k$ detectors and converting  their votes to probabilities such that their sum is equal to 1.

\begin{algorithm}[ht]
\caption{MSAD inference}
\label{alg:inference}
\begin{algorithmic}[1]
    \Function{Run\_Inference}{$T$, $M$, $k$, $c$}
        \State \textbf{Input:} 
        \State $T$ - Input time series
        \State $M$ - Model selector
        \State $k$ - Number of detectors to combine
        \State $c$ - Combination method
        \State \textbf{Output:} Final anomaly score

        \Comment{\textit{\textcolor{blue}{Model prediction per window}}}
        \State $W \gets $ \Call{Segment}{$T$, $M$.input\_size}
        \State $P \gets $ \Call{Predict}{$M$, $W$}
        
        \Comment{\textit{\textcolor{blue}{Average Strategy}}}
        \If{$c = $ ``average''}
            \State $\bar{P} \gets$ \Call{Mean}{$P$}
            \State \Call{Keep\_Top\_K}{$\bar{P}$, $k$}
            \State $w \gets$ \Call{Normalize}{$\bar{P}$}
        
        \Comment{\textit{\textcolor{blue}{Vote Strategy}}}
        \ElsIf{$c = $ ``vote''}
            \State $\bar{V} \gets$ \Call{Count\_Votes}{$P$}
            \State \Call{Keep\_Top\_K}{$\bar{V}$, $k$}
            \State $w \gets$ \Call{normalize}{$\bar{V}$}
        \EndIf

        \Comment{\textit{\textcolor{blue}{Score computation}}}
        \State $detectors \gets$ \Call{Select\_Detectors}{$w$}
        \State $S_T \gets$ \Call{Run\_Detectors}{$T$, $detectors$}
        \State $\bar{S_T} \gets$ \Call{Weighted\_Average}{$S_T$, $w$}
        
        \State \Return $final\_score$
    \EndFunction
\end{algorithmic}
\end{algorithm}

Algorithm~\ref{alg:inference} summarizes the inference phase of our proposed pipeline MSAD. Overall, our pipeline takes four inputs: the time series data $T$, the model selector $M$ trained in advance, the number of detectors $k$ to be combined, and the combination method $c$ (either \textit{average} or \textit{vote}).

Initially, the input time series $T$ is segmented into non-overlapping windows $W$ based on the input size $\ell$ of the model selector $M$, as described in Section~\ref{sec:preprocessing}. Subsequently, the model selector generates the probabilities $P$ for each segment in $W$. Depending on the chosen combination method $c$, the algorithm computes the weights for the detectors. The final combined anomaly score $\bar{S_T}$ for the time series is computed as $\bar{S_T} = \sum_{i=1}^{k} w_i \cdot S_{T,D_i}$. This approach ensures that the final score integrates information from multiple detectors, weighted according to their relevance, as predicted by the model selector.

\section{Experimental Evaluation}
\label{sec:exp}

We now describe in detail our experimental analysis. For additional information, we make all our material publicly available online~\cite{ourcode} and provide an interactive WebApp~\cite{ourwebsite} for navigating and exploring the experimental results.

\subsection{Experimental Setup and Settings}
\label{exp:setup}

\noindent \textbf{Technical setup: }
We implemented the deep learning-based model selection methods in Python 3.5 using the PyTorch library~\cite{NEURIPS2019_bdbca288}. For the feature-based approach, we used the TSFresh~\cite{CHRIST201872} and scikit-learn~\cite{JMLR:v12:pedregosa11a} libraries. We then used sktime~\cite{Lning2019sktimeAU} for the $Rocket$ algorithm implementation. For the AD methods, we used the implementation provided in the TSB-UAD benchmark~\cite{10.14778/3529337.3529354}. The evaluation was conducted on a server with Intel Core i7-8750H CPU 2.20GHz x 12, with 31.3GB RAM, and Quadro P1000/PCIe/SSE2 GPU with 4.2GB RAM, and on Jean Zay cluster with Nvidia Tesla V100 SXM2 GPU with 32 GB RAM.

\noindent \textbf{Datasets: }
For our evaluation purposes, we use the public datasets identified in the TSB-UAD benchmark~\cite{10.14778/3529337.3529354}. \revthree{The benchmark comprises 16 datasets from various domains as described in Table~\ref{SymbolTable}. Each dataset contains multiple time series with point-level anomaly labels, resulting in over 1980 distinct time series in total that we use in our experiments.}

\revthree{For our \textbf{in-distribution} experiments, we divide the benchmark into training, validation, and test sets. The results presented in this section come exclusively from the test set, which contains 497 time series that the models have not seen during training.} 

\revthree{For our \textbf{out-of-distribution} (OOD) experiments, we use the leave-one-out approach. Each model is trained on all datasets except one, which we later use for evaluation (repeated for all datasets). This ensures that the entire domain of the held-out dataset is unknown to the model, simulating \textit{out-of-distribution} scenarios that test the transfer learning capabilities of the models.}

\noindent \textbf{Anomaly Detection Methods: }
For the experimental evaluation, we select 12 different AD methods, summarized in Table~\ref{SymbolTable}. Out of these, 8 are fully unsupervised (i.e., they require no prior information on the anomalies to be detected): IForest, IForest1, LOF, MP, NormA, PCA, HBOS, and POLY. The remaining 4 methods are semi-supervised (i.e., they require some information related to normal behaviors), namely, OCSVM, AE, LSTM-AD, and CNN. For all these anomaly detection baselines, we set the parameter as described in the TSB-UAD benchmark~\cite{10.14778/3529337.3529354}.

\noindent \textbf{Method Selection baselines: }
We then consider the method selection baseline described in Section~\ref{sec:proposed} and summarized in Table~\ref{SymbolTable}. We first consider {\it feature-based} methods, that extract features using TSFresh~\cite{CHRIST201872} library to select the correct AD method. We then consider Rocket, state-of-the-art {\it time series classifier}. We also include two types of deep learning classifiers; (i) {\it Convolutional-based neural networks} and (ii) {\it Transformer-based neural networks}. Table~\ref{SymbolTable} summarizes the different model selection methods (i.e., classifiers). In total, we consider 16 methods, trained with window lengths $\ell$ equal to 16, 32, 64, 128, 256, 512, 768, and 1024. In total, we trained 128 models. In the following section, we refer to a model $M$ trained using a window length $\ell$ as $M$-$\ell$.

\noindent \textbf{Parameter settings: }
We use the same 70/30 split of the benchmark for all the classification models. Therefore, we can compare models trained on the same training set and evaluated on the same set of time series. Then, for the feature-based methods, we set the hyperparameters of the models based on the default parameters of scikit-learn. Moreover, for Rocket, we use 10000 kernels to extract the features and the logistic regression with stochastic gradient descent (computed in batches) for the classification step. Finally, for Convolutional and Transformer-based methods, we use a learning rate of $10^{-5}$, with a batch size of 256 and an early stopping strategy with a maximum of 50 epochs without improvement. Moreover, we use the weighted cross-entropy loss and set the maximum number of epochs to 10,000 (with a training time limit of 20 hours). \revthree{We use the default hyperparameters for all classifiers to ensure fairness and scalability, as hyperparameter tuning 128 different model configurations would not be computationally feasible. While these settings may not provide the best possible results for every classifier, they allow for a reasonable baseline. Our goal in this study is to assess the relative performance across model families and input settings rather than to optimize individual models.}

\noindent \textbf{Evaluation measures: }
We finally use four evaluation measures, summarized in Table~\ref{SymbolTable}. For model selection accuracy, we use the classification accuracy (i.e., the number of anomaly detectors correctly selected divided by the total number of time series). For anomaly detection accuracy, we use both AUC-PR~\cite{10.1145/1143844.1143874} and VUS-PR~\cite{10.14778/3551793.3551830} (with a buffer length equal to 10 points). For execution time, we measure the {\it training time} (i.e., the time required to train a model selection algorithm), the {\it selection time} (i.e., the time a model selection approach needs to predict which detector to use), and the {\it detection time} (i.e., the time required to predict which detector to use, and to execute it).  \revone{We focus on threshold-independent evaluation measures (AUC-PR and VUS-PR), as they provide a more robust assessment of performance, but we also compute 14 evaluation measures in total, including threshold-dependent ones, in our public repository to support practical applications.}

\begin{figure*}
    \centering
    \includegraphics[width=0.93\linewidth]{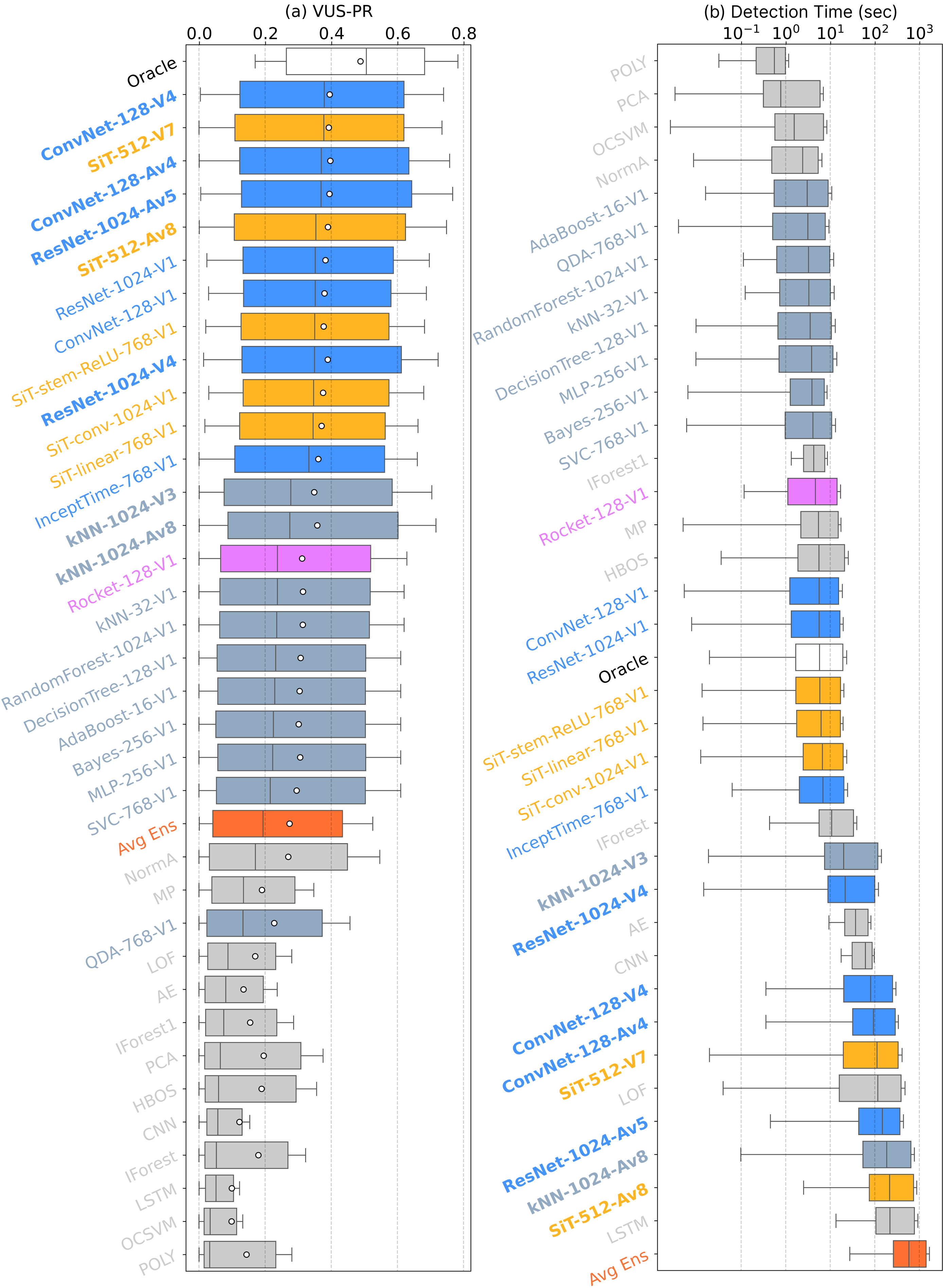}
        \caption{VUS-PR and Detection time (seconds) for all model selection approaches (showing only the window length and $k$ that maximize VUS-PR for each model) over a test set of 497 series from TSB-UAD. The methods are sorted: the most accurate methods are at the top (a); the fastest methods are at the top (b)}.
        \label{fig:overall_res}
\end{figure*}

\subsection{Overall Evaluation}
\label{exp:overalleval}

We first evaluate accuracy (classification and anomaly detection) and execution time for all model selection methods over the entire benchmark. We split the benchmark into a train and test set with $1404$ and $497$ time series, respectively. Both sets contain time series from all datasets. Thus, the models have examples of all available domains. In Section~\ref{exp:sup2unsup}, we evaluate the performance of the models when applied to unseen (i.e., not used in the training set) datasets.

\subsubsection{\textbf{Accuracy Evaluation}}

We first analyze the accuracy of all model selection methods (using all window lengths) and compare them to the Oracle, the Averaging Ensemble method (Avg Ensemble), and the AD methods in the TSB-UAD benchmark.

\begin{figure*}
    \centering
    \includegraphics[width=0.8\linewidth]{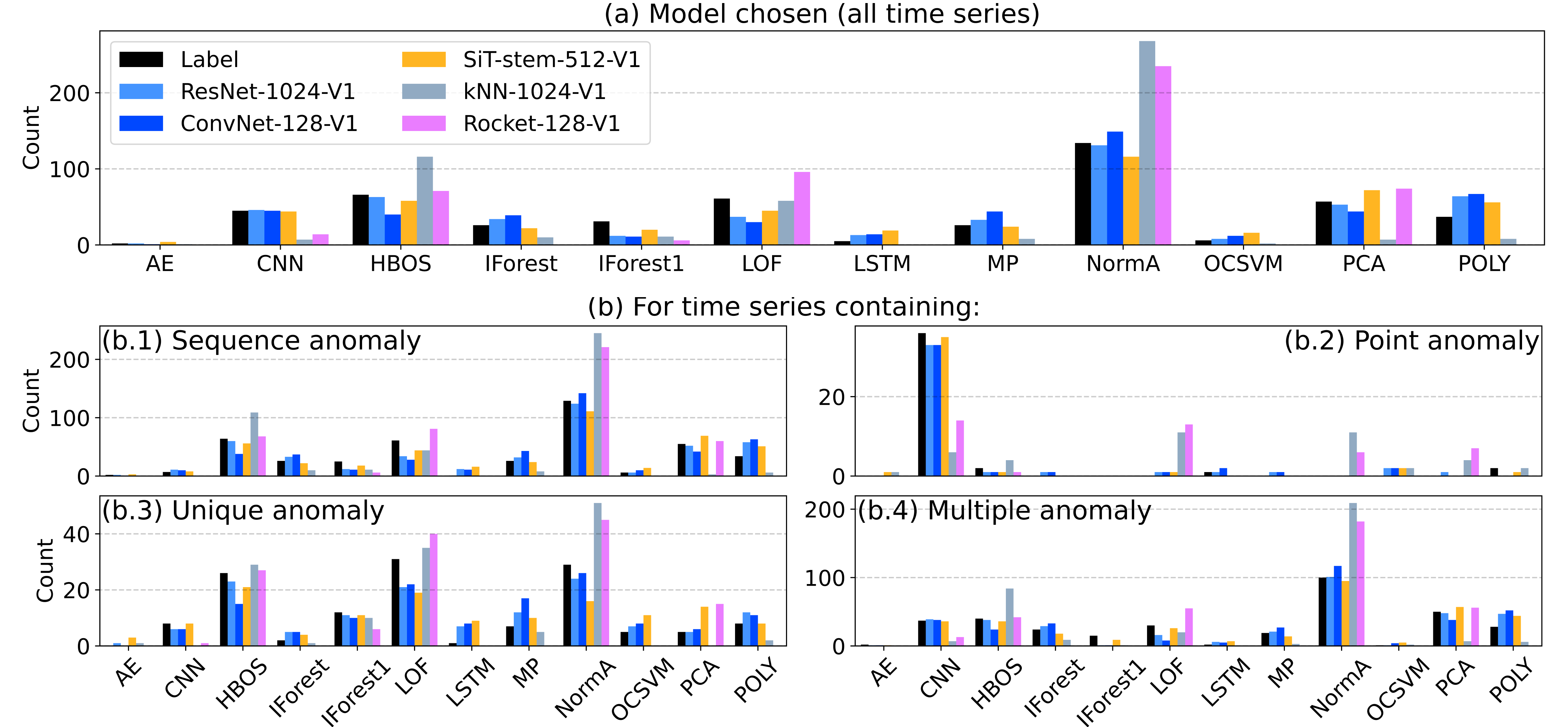}
        \caption{\revthree{Distribution of the selected models for five models (the best for each category) compared to the distribution of the labels (in black). Comparisons for time series containing (b) sequence vs. point anomalies, and (c) unique vs. multiple anomalies.}}
        \label{fig:classif_distrib}
\end{figure*}

Figure~\ref{fig:overall_res} (a) depicts the overall VUS-PR over the entire TSB-UAD benchmark (i.e., each box-plot corresponds to 497 accuracy values for the 497 time series into the test set). The Convolutional-based approaches are in dark blue, the Transformer-based approaches are in yellow, the Feature-based approaches are in light blue, $Rocket$ models are in violet, and the AD methods of the TSB-UAD benchmark are in light grey. The oracle is the top box plot (in white), and the Avg Ensemble is the orange box plot. The box plots are sorted based on the median value (the mean accuracy of each model is also displayed as a white circle). \revthree{In total, we compare 234 models on 497 time series, comprising 128 model selectors that were trained from scratch, 12 individual AD methods from the TSB-UAD benchmark, and 2 baselines, namely the Oracle and Averaging Ensemble. Out of the 128 trained models, the 4 top-performing model selectors were further tested for combining multiple detectors. In Figure~\ref{fig:overall_res}, we depict only the models with the window length that leads to the best VUS-PR for visual clarity.} The method used to combine probabilities and obtain the final weights is denoted by 'V' for vote and 'Av' for average. The subsequent number represents the value of \(k\), i.e., the number of detectors whose scores were combined for the final result. \revthree{For example, the model at the top of Figure~\ref{fig:overall_res} (a) named $ConvNet$-$128$-$V4$ refers to a Convolutional-based model selector that takes as input subsequences of length 128, and uses the vote method to combine the probabilities of the top 4 predicted detectors.}

First, almost all model selection methods outperform the existing AD methods. We also see that most model selection methods outperform the Avg Ensemble approach. Thus, we can conclude that model selection using time series classifiers significantly improves the state-of-the-art methods.

First, almost all model selection methods outperform the existing AD methods. We also see that most model selection methods outperform the Avg Ensemble approach. Thus, we can conclude that model selection using time series classifiers significantly improves the state-of-the-art methods. \revthree{However, we also observe that model selectors exhibit higher variance than individual anomaly detectors. We further discuss this at the end of this section.}

\revthree{More interestingly, we observe a partition in the ranking of the methods. First, Convolutional and Transformer-based approaches produce equivalent accuracy values and represent the top-48 methods (Note that not all models are shown here; in total, we evaluated 234 models)}. However, whereas all the Convolutional-based methods are in the top-48, a few of the Transformer-based approaches are further away in the ranking. Moreover, the first non-deep learning method is $Rocket$-$128$-$V1$ (ranked 49), followed closely by $kNN$ models. We also observe that the $Rocket$ approaches are very spread across the ranking ($Rocket$-$128$-$V1$ is ranked 50, and $Rocket$-$16$-$V1$ is ranked 124). This implies that the choice of window length strongly impacts accuracy. Overall, the best selection model is over 2 times more accurate than the best AD method in TSB-UAD.

Additionally, we discover that combining detectors (i.e., model selectors with \( k > 1 \)) \revtwo{yields slightly better results in the in-distribution setting (we will later demonstrate that the results are greatly improved in the out-of-distribution (OOD) setting)}. Almost all models that combine multiple detectors outperform their single-detector equivalents. Overall, the best selection model that combines detectors, i.e. $ConvNet$-$128$-$V4$ is $7.5\%$ more accurate than the best selection model with $k = 1$, i.e., $ResNet$-$1024$-$V1$.

Then, we also note that all model selection methods are significantly less accurate than the Oracle. For example, in Figure~\ref{fig:overall_res}(a), there is a gap of $0.12$ VUS-PR between the Oracle and the best model selection method, indicating substantial room for improvement. \revthree{Moreover, all model selection methods exhibit high variance in accuracy, as shown by the box plots in Figure~\ref{fig:overall_res}(a), including the Oracle, which is the theoretically perfect selector. This is primarily due to the presence of particularly difficult time series for which none of the available detectors perform well. As a result, even perfect model selectors cannot guarantee high performance across all cases. Making model selection more stable and robust remains an important challenge for future work. Despite this, model selectors consistently achieve strong performance on a large subset of time series, making them a more effective and flexible solution overall compared to individual detectors that often underperform consistently. That said, this trend does not hold in the OOD setting, as discussed in Section~\ref{exp:sup2unsup}. There, the $SiT$-$512$ model selector, when combining multiple detectors, almost guarantees equal to or better performance than the best individual AD method.}

\begin{figure}
    \centering
    \includegraphics[width=\linewidth]{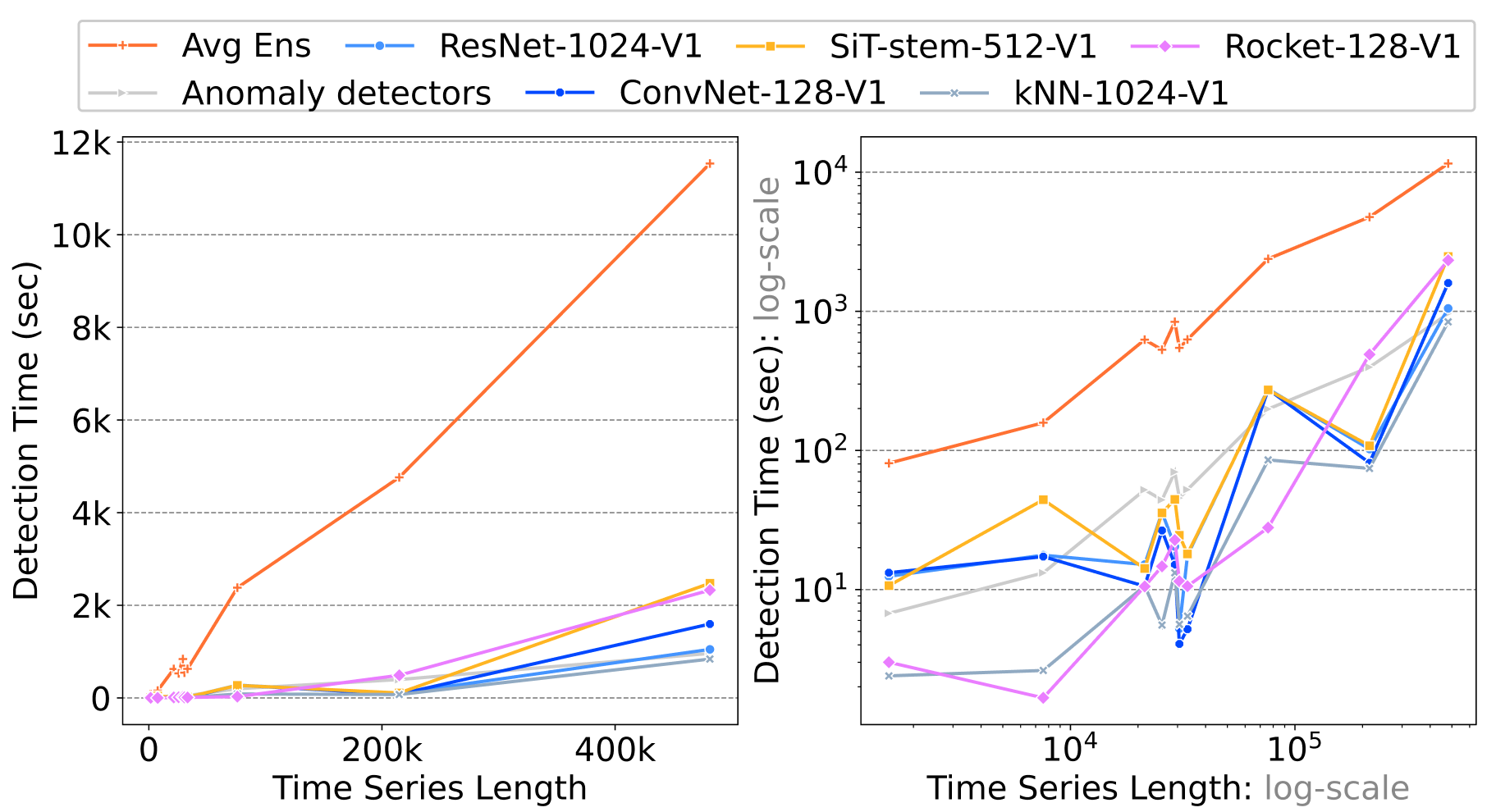}
        \caption{Execution time vs. length of model selection methods.}
        \label{fig:scalability}
\end{figure}

\subsubsection{\textbf{Model selected distribution}}
\label{exp:distribution}

We then examine the predictions, i.e. the detectors, selected by the model selection approaches. In this section, we consider only $ResNet$-$1024$-$V1$, $ConvNet$-$128$-$V1$, $SiT$-$stem$-$512$-$V1$ (or $SiT$-$512$-$V1$ for brevity), $kNN$-$1024$-$V1$, and $Rocket$-$128$-$V1$. These are the top-performing, single-detector models (evaluated using either AUC-PR or VUS-PR), based on the analysis conducted in Section~\ref{exp:overalleval}. Additional information on AUC-PR evaluation is available on our website~\cite{ourwebsite}. The corresponding $Av1$ models yield nearly identical results and are therefore not displayed here, but they can be reviewed in the project's repository.

Figure~\ref{fig:classif_distrib} (a) depicts the distribution of the chosen detectors by the model selection approaches mentioned above for the entire TSB-UAD benchmark. The black bar corresponds to the true labels (i.e., the best detectors). \revone{This analysis provides insight into how well model selectors capture the underlying time series characteristics and the types of anomalies they are most suited to detect.} We observe from Figure~\ref{fig:classif_distrib} (a) that $kNN$-$1024$-$V1$ and $Rocket$-$128$-$V1$ are significantly overestimating the detector NormA (as well as LOF for $Rocket$-$128$-$V1$ and HBOS for $kNN$-$1024$-$V1$), whereas $ResNet$-$1024$-$V1$, $ConvNet$-$128$-$V1$, and $SiT$-$512$-$V1$ are matching the correct distribution of detectors (we observe a slight underestimation of LOF, IForest1 and an overestimation for POLY). \revone{Overall, the deep learning-based model selectors show significantly better alignment with the ground truth, compared to the $kNN$ and $Rocket$ models, which tend to over-predict the majority class.}

Moreover, we measure the prediction distribution differences for time series containing sequence anomalies (Figure~\ref{fig:classif_distrib} (b.1)) and point anomalies (Figure~\ref{fig:classif_distrib} (b.2)), and for time series containing only one anomaly (Figure~\ref{fig:classif_distrib} (b.3)) and multiple anomalies (Figure~\ref{fig:classif_distrib} (b.4)). \revone{This breakdown allows us to evaluate how each classifier generalizes across different types of anomalies and structural variations in the data.} We first observe that predictions of model selection methods are significantly different for time series with sequence and point anomalies. More specifically, $ResNet$-$1024$-$V1$, $ConvNet$-$128$-$V1$, and $SiT$-$512$-$V1$ are correctly selecting the method CNN, whereas $kNN$-$1024$-$V1$ and $Rocket$-$128$-$V1$ are over selecting LOF and NormA for time series containing point anomalies. \revone{CNN and Transformer-based selectors are more sensitive to local spike-like patterns typical of point anomalies, while $kNN$ and $Rocket$ are not able to accurately identify them.} However, for sequence anomaly, as it represents most of the TSB-UAD benchmark, the prediction distribution is similar to the one over the entire benchmark. Moreover, the correct predictions of $ResNet$-$1024$-$V1$, $ConvNet$-$128$-$V1$, and $SiT$-$512$-$V1$ for time series containing point anomalies are interesting, as this information is not provided in the training step. \revone{Therefore, the deep learning-based model selectors found discriminant features in the time series that indicate whether it might contain a point or a sequence anomaly.}

Finally, we measure the differences between the prediction distributions of model selection methods between time series containing unique and multiple anomalies. The true labels (black bars in Figure~\ref{fig:classif_distrib} (b.3) and (b.4)) indicate that, for unique anomalies, the best detectors are LOF, NormA, and HBOS, and for multiple anomalies, the best detector is NormA. We observe that all model selection approaches tend to prefer LOF, NormA, and HBOS for time series containing a unique anomaly. \revone{The latter shows that model selection methods can extract discriminant features that indicate if one time series is more likely to have multiple anomalies, without explicit supervision. However, in cases of multiple anomalies, $ResNet$, $ConvNet$, and $SiT$ correctly identify that HBOS and LOF do not perform as well as NormA, and adjust their predictions accordingly. This adaptive behavior is not observed in the $kNN$ and $Rocket$ models.}

\subsubsection{\textbf{Execution Time Evaluation}}

We now discuss the execution time of model selection methods. In this section, we focus only on the detection time (i.e., the number of seconds required by a method to predict which detectors to use and to run them). Figure~\ref{fig:overall_res} (b) depicts the detection time (on a log scale) for each method and detector in the TSB-UAD benchmark. We first observe that the Avg Ensemble, which requires running all detectors, is significantly slower than the rest. Then, all model selection methods are of the same order of magnitude as the detectors. Even models with $k > 1$ remain within the same order of magnitude as the anomaly detectors, although, as expected, they are noticeably slower than single-detector models. We also observe that all the deep learning methods are slower than the feature-based approaches, except for $kNN$-$1024$-$V3$ and $kNN$-$1024$-$Av8$ which are slower due to selecting more than one detector. This is surprising, as detection time mainly depends on the chosen detector. Overall, we conclude that method selection is the only viable solution that outperforms the existing AD methods and can be executed in the same order of magnitude of time. 

Finally, in Figure~\ref{fig:scalability}, we depict the scalability of single-detector model selection methods versus individual detectors and the Avg Ensemble approach as the time series length increases (the \textit{average} equivalents of the model selectors shown, yield identical results and are thus not depicted here). We observe that, on average, the execution time of model selection approaches increases similarly to the execution time of individual detectors when the time series length increases. We also observe that the time series length significantly impacts the Avg Ensemble approach execution time. The latter shows the scalability issue of the Avg Ensemble approach for very large time series.

\begin{figure}
    \centering
    \includegraphics[width=\linewidth]{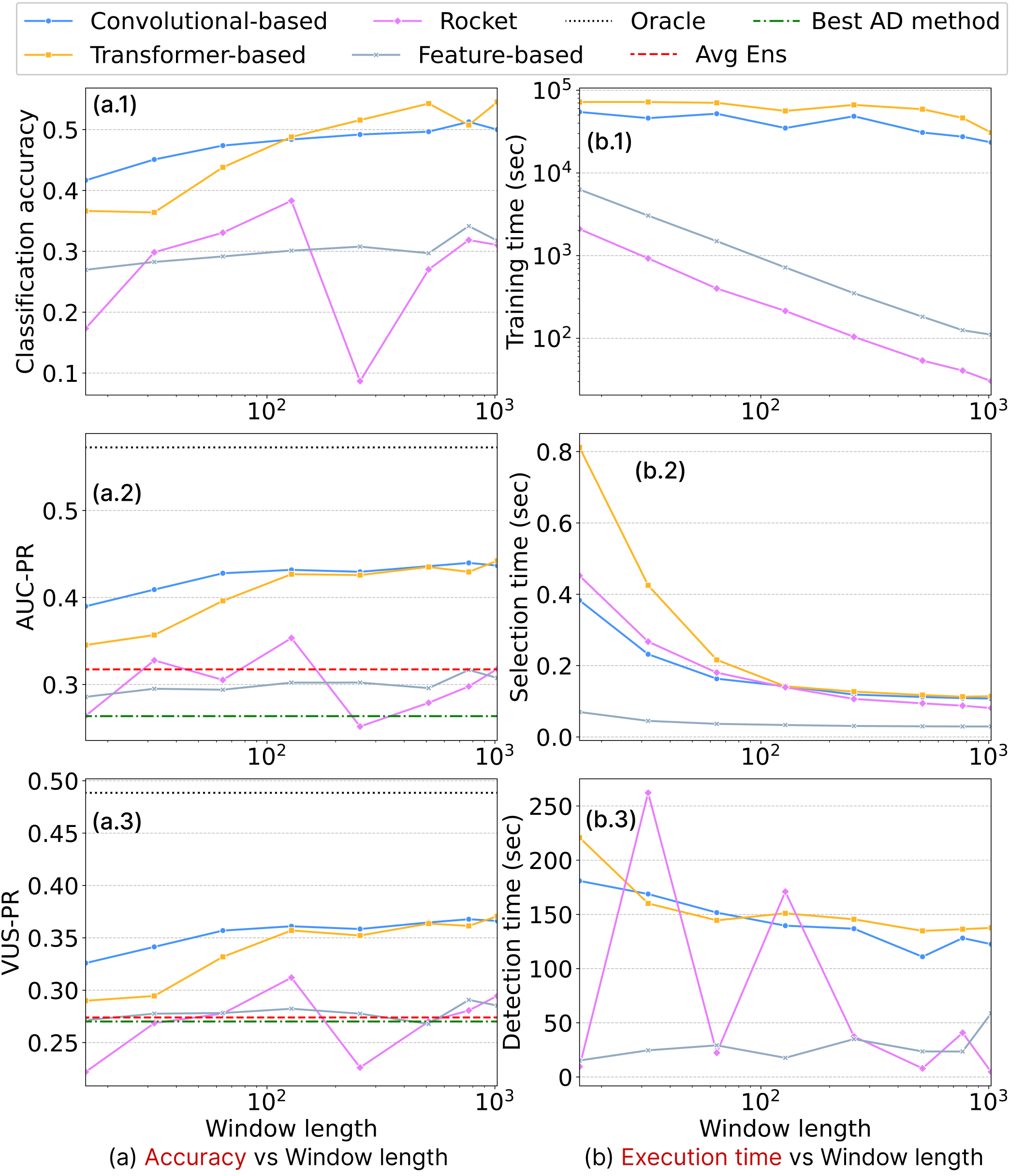}
        \caption{(a) Accuracy ((a.1) classification accuracy, (a.2) VUS-PR and (a.3) AUC-PR) and (b) execution time ((b.1) training time, (b.2) selection time and (b.3) detection time) versus window length $\ell$.}
        \label{fig:lengthinfl}
\end{figure}

\subsection{Influence of the Window Length}
\label{exp:windowlength}

\begin{figure*}
    \centering
    \includegraphics[width=0.8\linewidth]{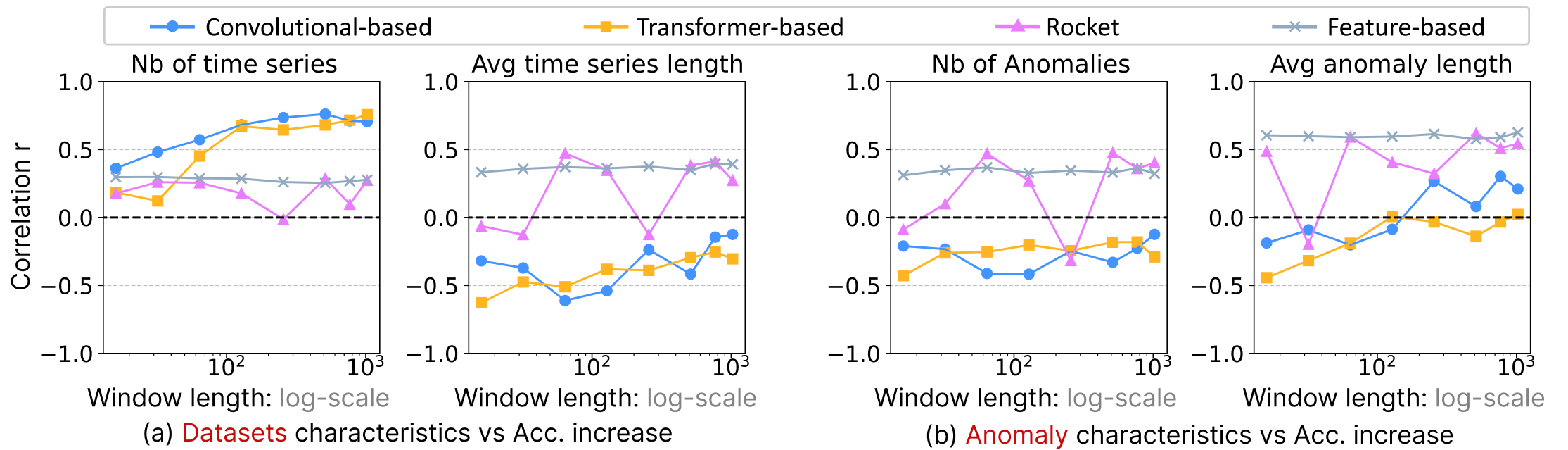}
    \caption{Correlation between accuracy and time series characteristics vs. the window length used to train the model selection.}
    \label{fig:infl_charac}
\end{figure*}

In this section, we analyze the influence of the window length on classification accuracy (Figure~\ref{fig:lengthinfl} (a.1)), anomaly detection accuracy (Figure~\ref{fig:lengthinfl} (a.2) and (a.3)) and execution time (Figure~\ref{fig:lengthinfl} (b)). We perform the analysis per group of methods (i.e., averages for Convolutional, Transformer, Rocket, and Feature-based methods), focusing exclusively on single-detector models that use the \textit{vote} combination method.

We first observe in Figure~\ref{fig:lengthinfl} (a) that Convolutional-based and Transformer-based methods outperform the best AD methods (green dashed line in Figure~\ref{fig:lengthinfl} (a.2) and (a.3)), the Avg Ensemble approach (orange dotted line in Figure~\ref{fig:lengthinfl} (a.2) and (a.3)), $Rocket$ and Feature-based methods, whatever the length used with regard to the classification accuracy, VUS-PR, and AUC-PR. \revone{Deep learning-based model selectors are more effective at capturing time series characteristics that are relevant for model selection, especially with longer windows lengths that allow them to better observe structure and trends.} We also note that Transformer-based approaches are less accurate for shorter lengths (less than 100 points), whereas the accuracy of Convolutional-based approaches is stable regardless of the window length. \revone{This difference likely reflects that Transformer models require more context to build meaningful representations, while Convolutional models can extract local patterns even from short windows.} Overall, Transformer and Convolutional-based approaches converge to the same anomaly detection accuracy (both for VUS-PR and AUC-PR) when the window length increases.

\begin{figure*}
    \centering
    \includegraphics[width=0.86\linewidth]{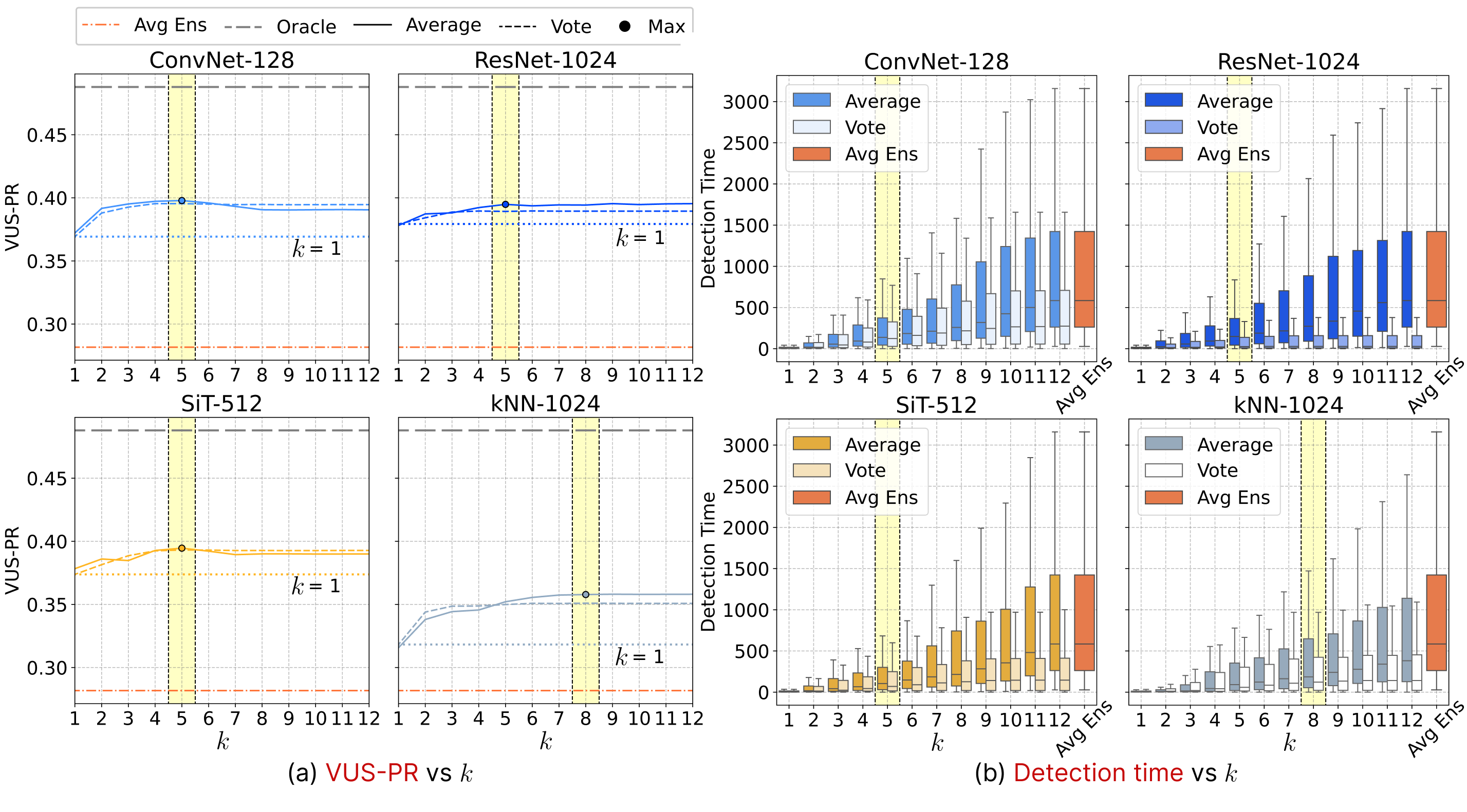}
        \caption{\revthree{VUS-PR vs $k$, i.e. detectors selected (the 4 plots on the left), and Execution time vs $k$ (the 4 plots on the right). The analysis is performed per model. The yellow rectangle highlights the k maximizing accuracy.}}
        \label{fig:k_comb_influence}
\end{figure*}

Furthermore, we observe that $Rocket$ and Feature-based approaches are both significantly faster to be trained than Convolutional and Transformer-based approaches (Figure~\ref{fig:lengthinfl} (b.1)). We make the same observation for selection time (Figure~\ref{fig:lengthinfl} (b.2)). For the detection time, we observe that $Rocket$ execution time is very unstable when compared to the other approaches. The latter means that the choice of length strongly impacts the model selection performed by Rocket, leading to very diverse selection and execution times. \revone{$Rocket$ may be more sensitive to how much context is available, and that its selection decisions vary more depending on the windowed input size.}

In the general case, we can make the following two statements: (i) A large window length results in faster selection time for the model selection process and better accuracy for Convolutional and Transformer-based approaches. \revone{Deep architectures benefit from having more input information, making them better suited to long, structured time series.} (ii) Feature-based approaches are significantly faster but less accurate than Convolutional-based and Transformer-based approaches, regardless of the window length used.

\subsection{Influence of Datasets and Anomaly Types}
\label{exp:datasets}

In this section, we evaluate the influence of datasets and anomaly characteristics on model selection accuracy. We perform the analysis per group of methods (i.e., average performances for Convolutional, Transformer, Rocket, and Feature-based methods), focusing exclusively on single-detector models that use the \textit{vote} combination method.

For this experiment, we evaluate the dataset and anomaly characteristics (i.e., the number of time series, the average length of the time series, the average number of anomalies and the average anomaly length). Figure~\ref{fig:infl_charac} depicts these characteristics (x-axis) versus the average increase of accuracy (VUS-PR of the model selection method subtracted by VUS-PR of the best AD method for each dataset) for each model selection method using a given window length. For instance, if a point (one model selection method on one dataset) is positive (above the black dotted line), then this model is more accurate on the corresponding dataset than the best AD method selected on this same dataset. We observe low correlations between dataset and anomaly characteristics (i.e., $-0.6<r<0.6$). Thus, we cannot conclude any statement on the impact of these characteristics and the model selection methods' performances. However, we can make the following observations.

First, Figure~\ref{fig:infl_charac} (a) shows that the number of time series is impacting more substantially Convolutional and Transformer-based approaches with large window lengths. For the average time series length, only Feature-based approaches are positively impacted. On the contrary, Convolutional and Transformer-based approaches are less accurate when the average time series length is increasing. \revone{These observations imply that deep learning-based selectors ($ConvNet$, $SiT$) benefit more from having a larger number of training examples rather than longer sequences. This is likely because additional time series provide more diverse information, whereas longer sequences may contain more repetitive patterns rather than new information.} In contrast, Feature-based approaches benefit from both more and large instances. 

\revone{Then, Figure~\ref{fig:infl_charac} (b) shows that Feature-based approach accuracy is increasing with the anomaly characteristics, whereas these characteristics negatively impact (or not at all) Convolutional and Transformer-based methods.} More specifically, we observe that Feature-based approaches (regardless of the window length) are more accurate with time series containing large anomalies, and Convolutional-based approaches are less accurate (irrespective of the window length) when the number of anomalies increases. \revone{This may suggest that more and larger anomalies have a stronger impact on the statistical features that Feature-based model selectors rely on, thereby aiding their performance. Time series with such anomalies are likely to stand out more in terms of statistical features that are being extracted, making them easier to classify correctly.}

We note that Rocket's correlation with the dataset and the anomaly characteristics is unstable. The latter is explained by the fact that the model prediction of $Rocket$ is sensitive to the window length (as described in Section~\ref{exp:windowlength}). Thus, it is impossible to conclude on Rocket's performances, datasets, and anomalies.

\subsection{Influence of $k$ and Combination Methods}
\label{exp:kandcombination}

In this section, we analyze the relationship between the anomaly detection accuracy of model selection methods (Figure~\ref{fig:k_comb_influence} (a)) and their execution time (Figure~\ref{fig:k_comb_influence} (b)) in relation to $k$, i.e. the number of detectors combined to produce the final anomaly score. 

First, in Figure~\ref{fig:k_comb_influence} (a), we observe that all models surpass the Avg Ensemble in anomaly detection accuracy, and that there is a consistent improvement when combining more than one detector (i.e., $k > 1$). However, this improvement is not constant, as the gains plateau after a certain point. More specifically, $ConvNet$-$128$, $ResNet$-$1024$, and $SiT$-$512$ achieve peak performance at $k = 5$, while $kNN$-$1024$ peaks at $k = 8$. Furthermore, for every model, the \textit{average} combination method yields better results than the \textit{vote} method. Although the difference between the two methods is not substantial, it is sufficient to prioritize the \textit{average} method when considering only the accuracy as main criteria.

Furthermore, not all models exhibit the same improvement in anomaly detection accuracy when combining multiple detectors. Specifically, $ConvNet$-$128$ shows an increase of 6.9\%, $ResNet$-$1024$ 4.5\%, $SiT$-$512$ 4.3\% and $kNN$-$1024$ 11.5\%. Interestingly, $kNN$-$1024$ benefits the most when combining detectors, while overall $ConvNet$-$128$ achieves the highest VUS-PR. Although combining detectors is promising, a gap remains when compared to the performance of the Oracle.

In Figure~\ref{fig:k_comb_influence} (b), we observe that the \textit{Vote} method is consistently faster as $k$ increases compared to \textit{Average}. This occurs because, although the model is instructed to use, for example, $k=9$, the \textit{Vote} method frequently selects fewer detectors (i.e., $k < 9$). This happens because some detectors are simply never selected across the windows of a time series, and thus can not contribute to the final result. In simple terms, the actual $k$ in the \textit{Vote} method is often smaller than the predefined $k$. This presents an opportunity to design a model selection approach that combines multiple detectors without requiring the parameter $k$, as the \textit{Vote} method naturally converges to a value of $k$. Importantly, while there is a difference in anomaly detection accuracy between the two combination methods, it is not substantial enough to disregard this concept from future research.

Finally, we observe a consistent increase in detection time across models as $k$ increases. However, the highest-performing models (highlighted in yellow boxes in Figure~\ref{fig:k_comb_influence} (b)) remain significantly faster than the Avg Ensemble, highlighting the value of running a subset of detectors for anomaly detection rather than all of them. As a rule of thumb, and based on the results described above, $k=5$ can be set as a default parameter.

\subsection{Detection vs Classification Accuracy}
\label{exp:detectionvsclass}

In this section, we analyze the relationship between the model selection methods' classification accuracy and the resulting anomaly detection accuracy. In this experiment, we consider VUS-PR as anomaly detection measure. For this experiment, we extend the definition of $Oracle$ (introduced in Section~\ref{sec:problem_def}) as follows:

\begin{definition}
    We define $Oracle_{k,j}$ as a hypothetical model selection method that has a classification accuracy of $k \in [0,1]$ and selects the $j^{th}$ best detector (among $m$ detectors) in cases of misclassification. Thus, $Oracle_{1,1}$ always selects the best detector, and $Oracle_{0,m}$ always selects the worst detector. Finally, we define $Oracle_{k,R}$ as the model selection method with a classification accuracy of $k \in [0,1]$ and that randomly selects a detector in misclassification cases.
    \end{definition}

\begin{figure}
    \centering
    \includegraphics[width=\linewidth]{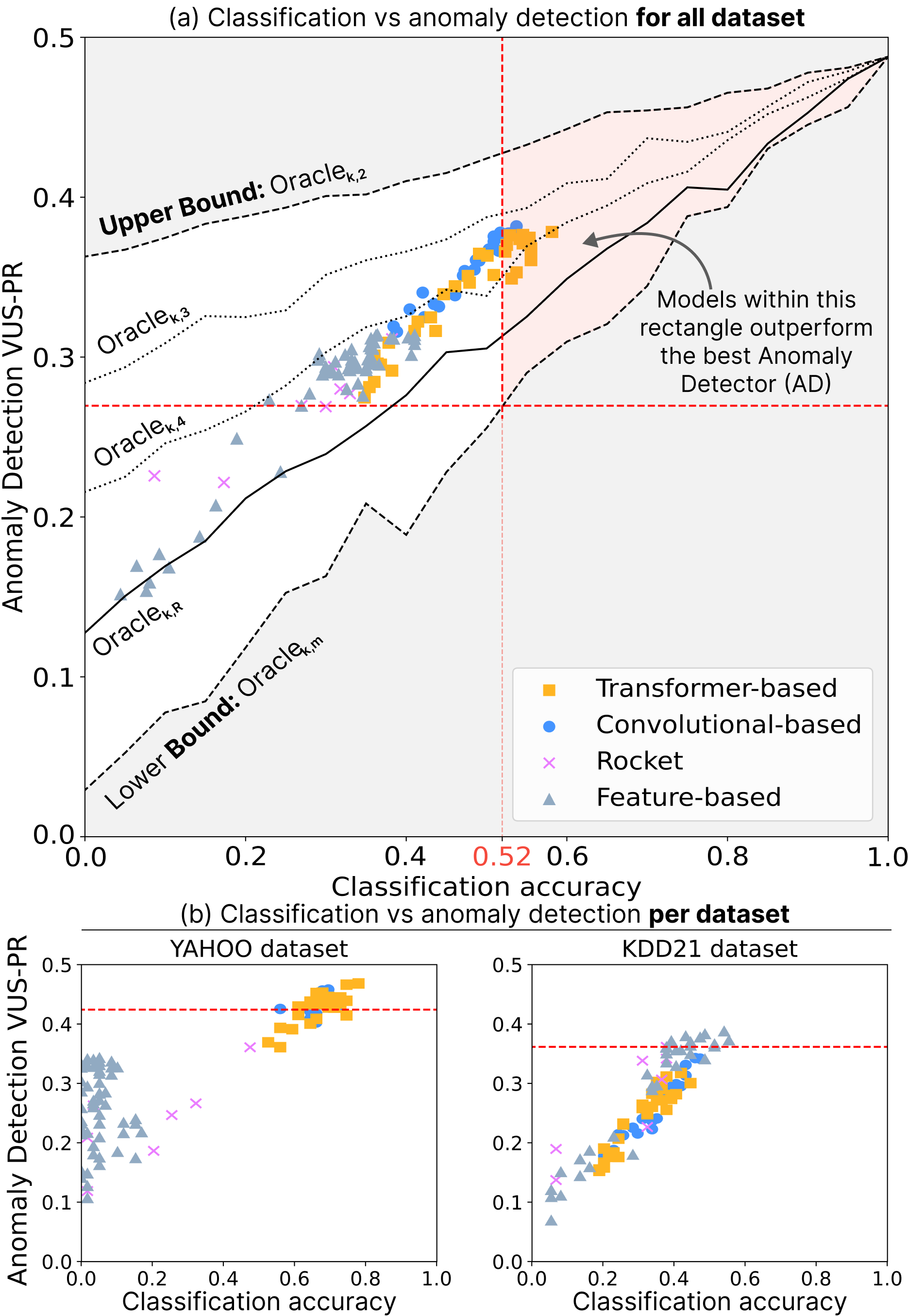}
        \caption{Classification vs. anomaly detection accuracy (VUS-PR) for (a) all datasets and (b) two specific datasets.}
        \label{fig:class_AD}
\end{figure}

Figure~\ref{fig:class_AD} depicts the latter comparison for all datasets (Figure~\ref{fig:class_AD} (a)), and two specific datasets (Figure~\ref{fig:class_AD} (b)). We first observe a strong correlation between classification accuracy and anomaly detection accuracy for each specific dataset and, on average, all datasets. However, methods belonging to different families (e.g., Feature-based or Transformer-based) are not performing the same. For instance, Figure~\ref{fig:class_AD} (a) shows that Feature-based approaches are not accurate for YAHOO but are the best models for KDD21. Overall, we observe that Convolutional and Transformer-based are more accurate in classification and anomaly detection (Figure~\ref{fig:class_AD}(b)). 

We also depict in Figure~\ref{fig:class_AD} (a) the lines corresponding to $Oracle_{k,2}$, $Oracle_{k,3}$, $Oracle_{k,4}$, $Oracle_{k,R}$, and $Oracle_{k,m}$. For a given classification accuracy, $k$, $Oracle_{k,2}$, and $Oracle_{k,m}$ correspond to the upper and lower bounds. The latter means that model selection approaches with a given classification accuracy will be within the previously mentioned upper and lower bounds for VUS-PR (i.e., in the gray zone in Figure~\ref{fig:class_AD} (a)). Thus, any model selection method that has a classification accuracy above 0.53 (intersection between the two dashed red lines) is better than the current best AD method in TSB-UAD (i.e., red dashed line in Figure~\ref{fig:class_AD} (b)). 
This is true only for a few Convolutional- and Transformer-based methods in our experiments.

Moreover, we compare the positions of the model selection methods with regard the $Oracle_{k,3}$, $Oracle_{k,4}$, and $Oracle_{k,R}$. We observe in Figure~\ref{fig:class_AD} (b) that almost all methods are above $Oracle_{k,R}$. The latter means that the model selection methods do not randomly select detectors when the wrong detector is selected. Moreover, most models follow the $Oracle_{k,4}$ line. The latter indicates that the models averagely select the third-best in case of misclassification. Finally, the observations discussed above demonstrate three important statements: (i) classification accuracy can be used as a proxy for anomaly detection accuracy, and without computing the anomaly detection accuracy, we can provide an anomaly detection accuracy lower and upper bounds; (ii) the gap between the best model selection and the top right corner of the gray zone shows that there is a significant margin for improvement for future work; (iii) the vertical gap between the models and the upper bound ($Oracle_{k,2}$) shows that there is an important margin of improvement in the prediction rank: a model with the same classification accuracy can gain up to $0.1$ VUS-PR if it better selects models.

\subsection{Out-of-Distribution Experiments}
\label{exp:sup2unsup}

\begin{figure}
    \centering
    \includegraphics[width=0.85\linewidth]{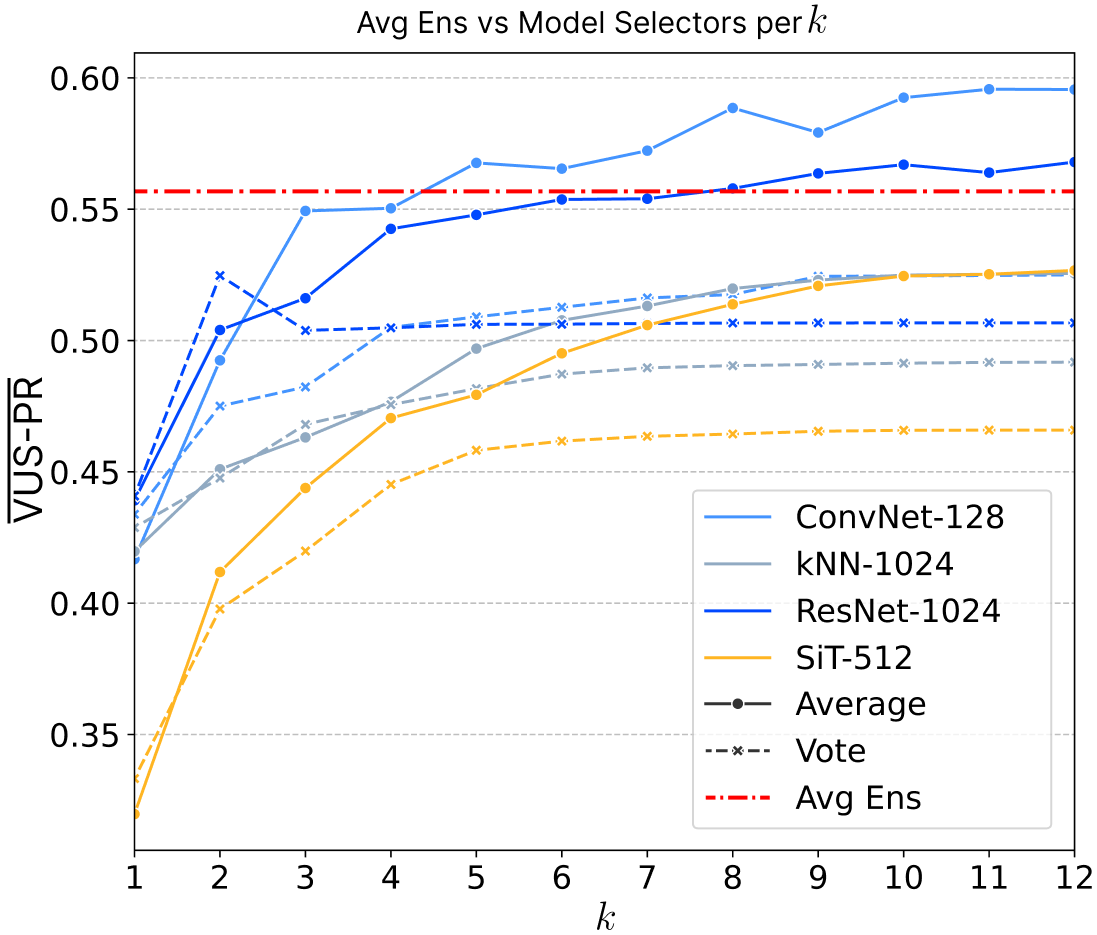}
        \caption{Out-of-distribution anomaly detection accuracy (normalized VUS-PR) versus $k$ per model selection method for \textit{Average} and \textit{Vote} combination methods.}
        \label{fig:k_vs_vus_unsup}
\end{figure}

\begin{figure}
    \centering
    \includegraphics[width=\linewidth]{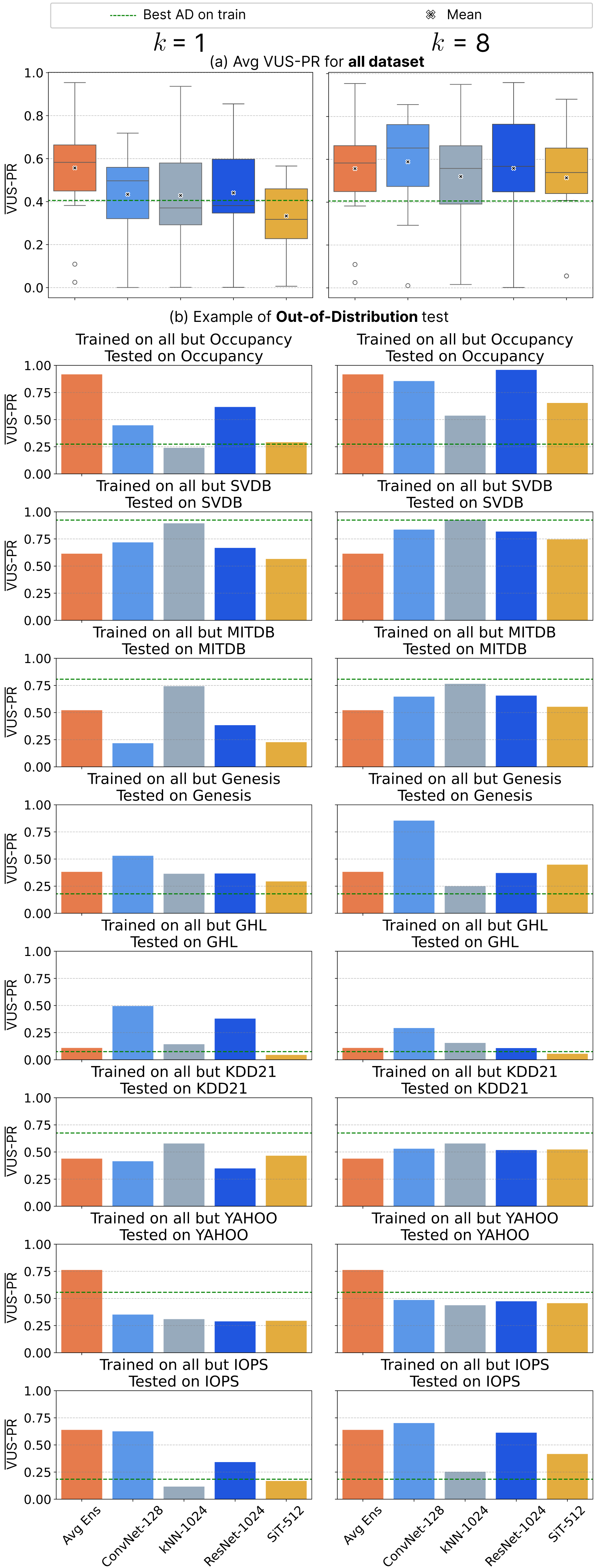}
        \caption{Out-of-distribution experiment. Comparison of model selection approaches (a) on average, and (b) per dataset, with (left) $k$ = 1 and (right) $k$ = 8}
        \label{fig:one_vs_all}
\end{figure}

Up to this point, we tested the performances of the model selection methods when trained on a subset of the benchmark with examples from all 16 datasets available. These results are interesting when we suppose that a user wants to analyze datasets similar to the one considered in the benchmark. In some cases, though, we may want to analyze time series that are not similar to any of those in the benchmark. Therefore, in this section, we measure the ability of the model selection methods to be used in an \revtwo{OOD} manner (i.e., used for datasets that are not similar to the one used in the training set). We run the following experiment. We train the model selection methods on 15 datasets (70\% of the time series for training and the other 30\% for validation), and we test on the remaining one. We try all 16 possible test partitions, and (for brevity) report 8 of these tests in Figure~\ref{fig:one_vs_all} (a). We only show the results for the best-performing model selection methods, namely, $ResNet$-$1024$, $ConvNet$-$128$, $SiT$-$512$, $kNN$-$1024$ and the Averaging ensemble for comparison. 

Figure~\ref{fig:k_vs_vus_unsup} shows the normalized VUS-PR, denoted $\overline{VUS\text{-}PR}$, for each model selection approach, across all values of $k$, and both combination methods, namely \textit{Average} and \textit{Vote}. A $\overline{VUS\text{-}PR}$ value of 1 corresponds to the Oracle's performance on each test, while 0 corresponds to the worst AD method. \revtwo{First, as in the in-distribution setting, we observe that the \textit{Average} combination method consistently outperforms \textit{Vote} when combining detectors. Unlike the in-distribution setting, the anomaly detection accuracy does not plateau after a certain $k$, but instead continuous to improve as $k$ increases. Notably, $ConvNet$ and $ResNet$ reach the performance of the Avg Ensemble at $k = 5$. Thus, combining detectors allows for a significant reduction in execution time, as models can achieve similar performance to the Avg Ensemble by running only 5 detectors instead of 12. Finally, we observe that the best-performing model is $ConvNet$-$128$, and that $SiT$-$512$ shows remarkable improvement, gaining more than 0.1 VUS-PR after combining only 4 detectors.}

Figure~\ref{fig:one_vs_all} (a) illustrates the normalized VUS-PR for all 16 tests. The figure shows that when using only single-detector models, we fail to match the performance of the Avg Ensemble in the \revtwo{OOD setting and barely surpass the accuracy of the best AD method, as measured on the train set (dotted green line in Figure~\ref{fig:one_vs_all} (a)). However, by combining detectors, we can both achieve the performance and reduce the execution time of the Avg Ensemble. As with the in-distribution setting, we observe that anomaly detection benefits from multiple-detector model selection in the OOD setting.} Another observation is that while $SiT$-$512$ does not achieve the highest performance, its box plot skewers suggest that it is the only model that can almost guarantee equal to or better performance than the best AD method on the train set.

Figure~\ref{fig:one_vs_all} (b) depicts the average accuracy for 8 out of the 16 tests (datasets excluded from the training set and used for testing). We observe very different results. First, for Electrocardiograms (SVDB), neither the model selection methods nor the Avg Ensemble outperforms the best AD method (selected on the training set). However, for various kinds of sensor data (GHL and Occupancy), model selection methods and the Avg Ensemble do outperform the best AD method. This difference can be explained by the fact that ECGs exhibit less diverse behaviors (i.e., repetitive normal patterns and similar anomalies) than other sensor data. Consequently, it is more likely to have one method that performs well on all ECG time series. This observation is supported by the fact that the performance of the best AD method closely matches that of the Oracle for SVDB. Interestingly, in the GHL dataset, combining detectors, i.e., $k > 1$, reduces the performance. This indicates the poor ranking produced by model selectors for this specific dataset, confirming the need for future research towards rank-based training and prediction for model selection (i.e., conclusion of Section~\ref{exp:detectionvsclass}). Finally, the use of $k > 1$ is critical to ensure performance similar to or better than Avg Ensemble, as seen in the Occupancy and Genesis datasets. Combining detectors to produce the final anomaly score is noticeably beneficial. 

Overall, we observe that: (i) combining multiple detectors (i.e., $k > 1$) is crucial in the \revtwo{out-of-distribution} setting to achieve performance comparable to the Avg Ensemble; (ii) the performance of model selection methods varies significantly across different types of time series data; (iii) classifiers as model selection can be used for TSAD, even though similar time series are not in the training set.

\section{Conclusion}
\label{sec:conclusions}

TSAD is a challenging problem and an important area of research with applications  in many scientific, societal, and industrial domains. Despite the multitude of solutions proposed in the literature, we observe that there exists no method that outperforms all others when measured on large heterogeneous benchmarks. Based on our experimental evaluation, we answer the questions of Section~\ref{sec:objective} as follows:

\begin{enumerate}
	\item \textbf{Classification as Model selection}: We observe that time series classification methods accurately select anomaly detection models. Overall, Transformer and Convolutional-based model selection methods outperform each individual detector. Nevertheless, there is a large gap between the best method and the $Oracle$, motivating future work toward that direction. \revtwo{An interesting next step could involve incorporating \textit{detector diversity} into the model selection process. We observe that certain detectors exhibit strong correlations in their anomaly score patterns, while others are largely uncorrelated. This suggests that combining diverse (i.e., less correlated) but accurate detectors could yield more informative anomaly scores than combining similar ones. Integrating such diversity-aware strategies while still preserving computational efficiency remains an open and promising challenge for future work.}
    \revone{\item \textbf{Single vs. multiple detectors}: We find that combining even a few detectors significantly improves performance. In the OOD case, combining detectors ($k > 1$) is necessary to outperform the \textit{Avg Ens}.}
	\item \textbf{Ensembling or selecting}: We observe that model selection is significantly more accurate than the Ensembling method. Moreover, in the in-distribution setting, $k=1$ is sufficient to significantly outperform Ensembling.
	\item \textbf{Features or Raw values}: We observe that raw-based methods are more accurate on average than feature-based approaches.
	\item \textbf{Out-Of-Distribution}: (1) and (3) hold. However, for (2), we observe that model selection with $k=1$ is not enough to reach the performance of the ensembling method when applied to time series very different from those in the training benchmark. Nevertheless, $k=5$ enables model selection to reach the accuracy of ensembling while reducing significantly the overall execution time. Finally, model selection with larger values of $k>5$ outperforms ensembling in the out-of-distribution setting, however the benefit in terms of execution time is rather limited. 
\end{enumerate}
The above observations point to promising directions for future work in AutoML frameworks that rely on model selection. As mentioned in Section~\ref{exp:detectionvsclass}, improving the rank prediction could significantly improve the anomaly detection accuracy. Moreover, model selection could be trained to choose the best compromise between accuracy and execution time, improving the overall inference time of model selection. 
\revthree{Finally, to support real-world adoption, we provide the following recommendations:
\begin{enumerate}
    \item Based on our experiments, we recommend using window sizes of at least 128. Values like 512 or 1024 are both effective, so users may choose based on the expected size of the patterns in their data.
    \item Deep learning model selectors, such as $ConvNet$-$128$ and $SiT$-$512$, perform consistently well in both in-distribution and OOD scenarios, making them reliable default choices.
    \item For combining predicted detectors, both voting and averaging are accurate in-distribution. However, in the OOD setting, voting provides a better trade-off between performance and efficiency.
    \item In the in-distribution case, selecting a single detector is often sufficient. In contrast, combining multiple detectors is crucial for OOD robustness, where even $k=5$ enables strong performance gains, while remaining computationally efficient.
\end{enumerate}
These insights aim to help practitioners apply model selection techniques effectively in practice and highlight the need for more robust strategies in time series anomaly detection. All our code is made available at \url{https://github.com/sylligardos/MSADv2}.}


\begin{acknowledgements}
Supported by EU Horizon projects AI4Europe ($101070000$), TwinODIS ($101160009$), DataGEMS ($101188416$) and RECITALS ($101168490$), by $Y \Pi AI \Theta A$ \& NextGenerationEU project HARSH ($Y\Pi3TA-0560901$) that is carried out within the framework of the National Recovery and Resilience Plan “Greece 2.0” with funding from the European Union – NextGenerationEU, and Cyberté (BPI-funded project). This work was granted access to the HPC resources of IDRIS under the allocation 2025-A0191012641 made by GENCI.
\end{acknowledgements}


\bibliographystyle{spmpsci}
\bibliography{sylli_bibliography}

\end{document}